\documentclass[twocolumn]{article}
\usepackage[utf8]{inputenc}
\usepackage[a4paper, total={7in, 8in}]{geometry}

\usepackage{framed,multirow}

\usepackage{amssymb}
\usepackage{latexsym}

\usepackage{url}
\usepackage{xcolor}

\definecolor{newcolor}{rgb}{.8,.349,.1}

\usepackage{natbib}
\usepackage{graphicx}
\usepackage{tabularx}
\usepackage{booktabs}
\usepackage{array}
\usepackage{multirow}
\usepackage{amsmath}
\usepackage{enumitem}
\usepackage{siunitx}
\usepackage[normalem]{ulem}
\usepackage{subcaption}

\usepackage{authblk}
\usepackage{mathtools}
\usepackage[switch]{lineno}

\begin{document}



\title{Disentangled representation learning in cardiac image analysis}

\author[1]{Agisilaos Chartsias}
\author[1]{Thomas Joyce}
\author[2,3]{Giorgos Papanastasiou}
\author[2,3]{Michelle Williams}
\author[2,3]{David Newby}
\author[4]{Rohan Dharmakumar}
\author[1,5]{Sotirios A. Tsaftaris}

\affil[1]{Institute for Digital Communications, School of Engineering, University of Edinburgh, West Mains Rd, Edinburgh EH9 3FB, UK}
\affil[2]{Edinburgh Imaging Facility QMRI, Edinburgh, EH16 4TJ, UK}
\affil[3]{Centre for Cardiovascular Science, Edinburgh, EH16 4TJ, UK}
\affil[4]{Cedars Sinai Medical Center Los Angeles CA, USA}
\affil[5]{The Alan Turing Institute, London, UK}

\date{}

\maketitle

\begin{abstract}
Typically, a medical image offers spatial information on the anatomy (and pathology) modulated by imaging specific characteristics. Many imaging modalities including Magnetic Resonance Imaging (MRI) and Computed Tomography (CT) can be interpreted in this way. We can venture further and consider that a medical image naturally factors into some spatial factors depicting anatomy and factors that denote the imaging characteristics. Here, we explicitly learn this decomposed (disentangled) representation of imaging data, focusing in particular on cardiac images. We propose Spatial Decomposition Network (SDNet), which factorises 2D medical images into spatial anatomical factors and non-spatial modality factors. 
We demonstrate that this high-level representation is ideally suited for several medical image analysis tasks, such as semi-supervised segmentation, multi-task segmentation and regression, and image-to-image synthesis.  Specifically, we show that our model can match the performance of fully supervised segmentation models, using only a fraction of the labelled images.  Critically, we show that our factorised representation also benefits from supervision obtained either when we use auxiliary tasks to train the model in a multi-task setting (e.g. regressing to known cardiac indices), or when aggregating multimodal data from different sources (e.g. pooling together MRI and CT data). To explore the properties of the learned factorisation, we perform latent-space arithmetic and show that we can synthesise CT from MR and vice versa, by swapping the modality factors. We also demonstrate that the factor holding image specific information can be used to predict the input modality with high accuracy. Code will be made available at \url{https://github.com/agis85/anatomy_modality_decomposition}.
\end{abstract}



\section{Introduction} \label{sec:introduction}

Learning good data representations is a long running goal of machine learning \citep{bengio2013representation}. In general, representations are considered ``good'' if they capture explanatory (discriminative) factors of the data, and are useful for the task(s) being considered. Learning good data representations for medical imaging tasks poses additional challenges, since the representation must lend itself to a range of medically useful tasks, and work across data from various image modalities.

Within deep learning research there has recently been a renewed focus on methods for learning so called ``disentangled'' representations, for example in \cite{higgins2016beta} and \cite{chen2016infogan}. A disentangled representation is one in which information is represented as a number of (independent) factors, with each factor corresponding to some meaningful aspect of the data \citep{bengio2013representation} (hence why sometimes encountered as factorised representations). 

Disentangled representations offer many benefits: for example, they ensure the preservation of information not directly related to the primary task, which would otherwise be discarded, whilst they also facilitate the use of only the relevant aspects of the data as input to later tasks. Furthermore, and importantly, they improve the interpretability of the learned features, since each factor captures a distinct attribute of the data, while also varying independently from the other factors.

\subsection{Motivation}

Disentangled representations have considerable potential in the analysis of medical data. In this paper we combine recent developments in disentangled representation learning with strong prior knowledge about medical image data: that it necessarily decomposes into an ``anatomy factor'' and a ``modality factor''. 

An anatomy factor that is explicitly spatial (represented as a multi-class semantic map) can maintain pixel-level correspondences with the input, and directly supports spatially equivariant tasks such as segmentation and registration. Most importantly, it also allows a meaningful representation of the anatomy that can be generalised to any modality. As we demonstrate below, a spatial anatomical representation is useful for various modality independent tasks, for example in extracting segmentations as well as in calculating cardiac functional indices. It also provides a suitable format for pooling information from various imaging modalities. 

The non-spatial modality factor captures global image modality information, specifying how the anatomy is rendered in the final image. Maintaining a representation of the modality characteristics allows, among other things, the ability to use data from different modalities.

Finally, the ability to learn this factorisation using a very limited number of labels is of considerable significance in medical image analysis, as labelling data is tedious and costly. 
Thus, it will be demonstrated that the proposed factorisation, in addition to being intuitive and interpretable, also leads to considerable performance improvements in segmentation tasks when using a very limited number of labelled images.

\begin{figure}[t]
\centering
\includegraphics[width=\linewidth]{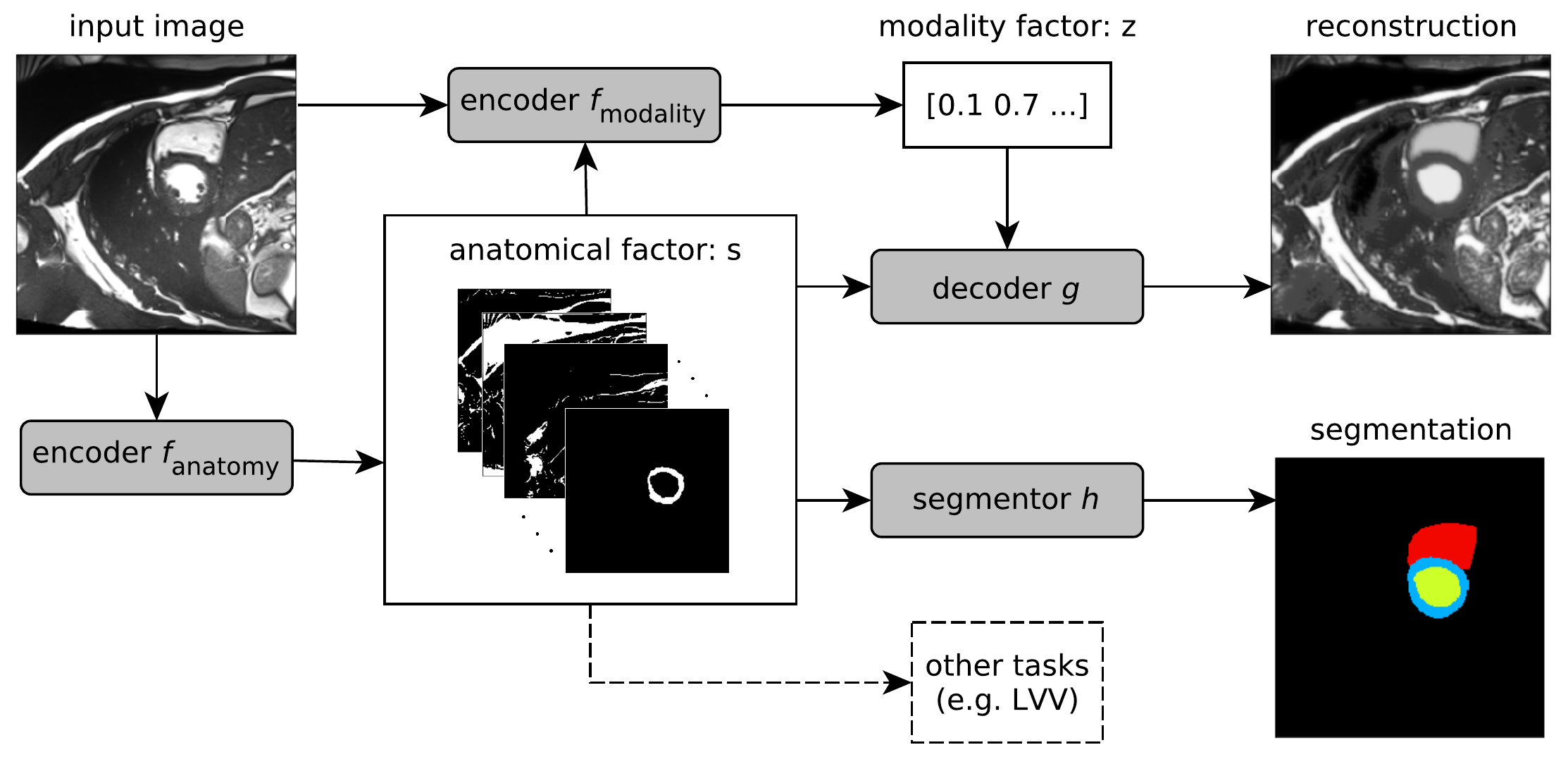}
\caption{A schematic overview of the proposed model. An input image is first encoded to a multi-channel spatial representation, the anatomical factor $s$, using an anatomy encoder $f_{anatomy}$. Then $s$ can be used as an input to a segmentation network $h$ to produce a multi-class segmentation mask, (or some other task specific network). The factor $s$ along with the input image are used by a modality encoder $f_{modality}$ to produce a latent vector $z$ representing the imaging modality. The two representations $s$ and $z$ are combined to reconstruct the input image through the decoder network $g$. \label{fig:simpleschematic}}
\end{figure}

\subsection{Overview of the proposed approach}

Learning a decomposition of data into a spatial content factor and a non-spatial style factor has been a focus of recent research in computer vision \citep{huang2018munit, lee2018diverse} with the aim being to achieve diversity in style transfer between domains. However, no consideration has been taken regarding the semantics and the precision of the spatial factor. This is crucial in medical analysis tasks in order to be able to extract quantifiable information directly from the spatial factor. Concurrently with these approaches, \cite{chartsias2018factorised} aimed to precisely address the need for interpretable semantics by explicitly enforcing the spatial factor to be a binary myocardial segmentation. However, since the spatial factor is a segmentation mask of only the myocardium, remaining anatomies must be encoded in the non-spatial factor, which violates the concept of explicit factorisation into anatomical and modality factors.

In this paper instead, we propose \textit{Spatial Decomposition Network} (SDNet), schematic shown in Figure~\ref{fig:simpleschematic}, that learns a disentangled representation of medical images consisting of a spatial map that semantically represents the anatomy, and a non-spatial latent vector containing image modality information. 

The anatomy is modelled as a multi-channel feature map, where each channel represents different anatomical substructures (e.g. myocardium, left and right ventricles). This spatial representation is categorical with each pixel necessarily belonging to exactly one channel. This strong restriction prevents the binary maps from encoding modality information, encouraging the anatomy factors to be modality-agnostic (invariant), and further promotes factorisation of the subject's anatomy into meaningful topological regions.

On the other hand, the non-spatial factor contains modality-specific information, in particular the distribution of intensities of the spatial regions. We encode the image intensities into a smooth latent space, using a Variational Autoencoder (VAE) loss, such that nearby values in this space correspond to neighbouring values in the intensity space.

Finally, since the representation should retain most of the required information about the input (albeit in two factors), image reconstructions are possible by combining both factors.

In the literature the term ``factor'' usually refers to either a single dimension of a latent representation, or a meaningful aspect of the data (i.e. a group of dimensions) that can vary independently from other aspects. Here we use factor in the second sense, and we thus learn a representation that consists of a (multi-dimensional) anatomy factor, and a (multi-dimensional) modality factor. Although the individual dimensions of the factors could be seen as (sub-)factors themselves, for clarity we will refer to them as dimensions throughout the paper.

\subsection{Contributions}

Our main contributions are as follows:
\begin{itemize}
    \item With the use of few segmentation labels and a reconstruction cost, we learn a multi-channel spatial representation of the anatomy. We specifically restrict this representation to be semantically meaningful by imposing that it is a discrete categorical variable, such that different channels represent different anatomical regions.
    
    \item We learn a modality representation using a VAE, which allows sampling in the modality space. This facilitates the decomposition, permits latent space arithmetic, and also allows us to use part of our network as a generative model to synthesise new images.
    
    \item We detail design choices, such as using Feature-wise Linear Modulation (FiLM) \citep{perez2017film} in the decoder, to ensure that the modality factors do not contain anatomical information, and prevent posterior collapse of the VAE.
    
    \item We demonstrate our method in a multi-class segmentation task, and on different datasets, and show that we maintain a good performance even when training with labelled images from only a single subject.
    
    \item We show that our semantic anatomical representation is useful for other anatomical tasks, such as inferring the Left Ventricular Volume (LVV). More critically, we show that we can also learn from such auxiliary tasks demonstrating the benefits of multi-task learning, whilst also improving the learned representation.
    
    \item Finally, we demonstrate that our method is suitable for multimodal learning (here multimodal refers to multiple modalities and not multiple modes in a statistical sense), where a single encoder is used with both MR and CT data, and show that information from additional modalities improves segmentation accuracy. 
\end{itemize}

In this paper we advance our preliminary work \citep{chartsias2018factorised} in the following aspects: 1) we learn a general anatomical representation useful for multi-task learning; 2) we perform multi-class segmentation (of multiple cardiac substructures); 3) we impose a structure in the imaging factor which follows a multi-dimensional Gaussian distribution, that allows sampling and improves generalisation;  4) we formulate the reconstruction process to use FiLM normalisation \citep{perez2017film}, instead of concatenating the two factors; and 5) we offer a series of experiments using four different datasets to show the capabilities and expressiveness of our representation. 

The rest of the paper is organised as follows: Section \ref{sec:related_work} reviews related literature in representation learning and segmentation. Then, Section \ref{sec:methods} describes our proposed approach. Sections \ref{sec:experiments} and \ref{sec:results} describe the setup and results of the experiments performed. Finally, Section \ref{sec:conclusion} concludes the manuscript.

\begin{figure*}[!t]
\centering
\includegraphics[width=\linewidth]{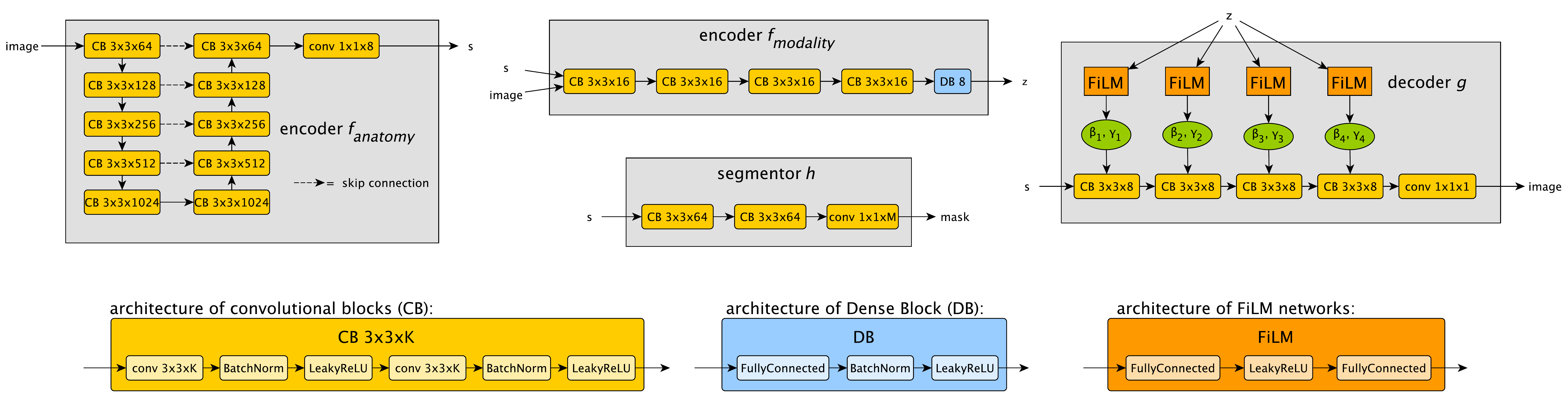}
\caption{The architectures of the four networks that make up SDNet. The anatomy encoder is a standard U-Net \citep{ronneberger2015u} that produces a spatial anatomical representation $s$. The modality encoder is a convolutional network (except for a fully connected final layer) that produces the modality representation $z$. The segmentor is a small fully convolutional network that produces the final segmentation prediction of a multi-class mask (with L classes) given $s$. Finally the decoder produces a reconstruction of the input image from $s$ with its output modulated by $z$ through FiLM normalisation~\citep{perez2017film}. The bottom of the figure details the components used throughout the four networks. The anatomical factor's channels parameter $C$, the modality factor's size $n_z$, and the number of segmentation classes $L$ depend on the specific task and are detailed in the main text.}
\label{fig:schematic}
\end{figure*}

\section{Related work} 
\label{sec:related_work}
Here we review previous work on disentangled representation learning, which is typically a focus of research on generative models (Section \ref{subsec:factorised}). We then review its application in domain adaptation, which is achieved by a factorisation of style and content  (Section \ref{subsec:stylecontent}). Finally, we review semi-supervised methods in medical imaging, as well as recent literature in cardiac segmentation, since they are related to the application domain of our method (Sections \ref{subsec:semisupervised} and \ref{subsec:cardiac_segmentation}).

\subsection{Factorised representation learning} 
\label{subsec:factorised}
Interest in learning independent factors of variation of data distributions is growing. 
Several variations of VAE~\citep{kingma2013auto, rezende2014stochastic} and Generative Adversarial Networks (GAN)~\citep{goodfellow2014generative} have been proposed to achieve such a factorisation. For example $\beta$-VAE \citep{higgins2016beta} adds a hyperparameter $\beta$ to the KL-divergence constraint, whilst Factor-VAE \citep{kim2018disentangling} boosts disentanglement by encouraging independence between the marginal distributions. On the other hand, using GANs, InfoGAN \citep{chen2016infogan} maximises the mutual information between the generated image and a latent factor using adversarial training, and SD-GAN \citep{donahue2017semantically} generates images with a common identity and varying style.
Combinations of VAE and GANs have also been proposed, for example by \cite{mathieu2016disentangling} and \cite{szabo2017challenges}. Both learn two continuous factors: one dataset specific factor, in their case class labels, and one factor for the remaining information. To promote independence of the factors and prevent a degenerate condition where the decoder uses only one of the two factors, mixing techniques have also been proposed \citep{hu2017disentangling}. 
These ideas also begin to see use in medical image analysis: \cite{Biffi2018_LearningInterpretable} apply VAE to learn a latent space of 3D cardiac segmentations to train a model of cardiac shapes useful for disease diagnosis. Learning factorised features is also used to distinguish between (learned) features specific to a modality from those shared across modalities \citep{fidon2017scalable}. However, their aim is combining information from multimodal images and not learning semantically meaningful representations.

These methods rely on learning representations in the form of latent vectors. Our method is similar in concept with \cite{mathieu2016disentangling} and \cite{szabo2017challenges}, which both learn a factorisation into known and other residual factors. However, we constrain the known factor to be spatial, since this is naturally related to the anatomy of medical images.

\subsection{Style and content disentanglement} \label{subsec:stylecontent}

There is a connection between our task and style transfer (in medical image analysis known as modality transformation or synthesis): the task of rendering one image in the ``style'' of another.
Classic style transfer methods do not explicitly model the style of the output image and therefore suffer from style ambiguity, where many outputs correspond to the same style. In order to address this ``many to one'' problem, a number of models have recently appeared that include an additional latent variable capturing image style. For example, colouring a sketch may result in different images (depending on the colours chosen) thus, in addition to the sketch itself, a vector parameterising the colour choices is also given as input \citep{zhu2017toward}.

Our approach here can be seen as similar to a disentanglement of an image into ``style'' and ``content'' \citep{gatys2016image, azadi2018multi}, where we represent content (i.e. in our case the underlying anatomy) spatially.
Similar to our approach, there have been recent disentanglement models that also use vector and spatial representations for the style and content respectively \citep{almahairi2018augmented, huang2018munit, lee2018diverse}. Furthermore, \cite{esser2018variational} expressed content as a shape estimation (using an edge extractor and a pose estimator) and combined it with style obtained from a VAE. The intricacies of medical images differentiate us by necessitating the expression of the spatial content factor as categorical in order to produce a semantically meaningful (interpretable) representation of the anatomy, which cannot be estimated and rather needs to be learned from the data. This discretisation of the spatial factor also prevents the spatial representation from being associated with a particular medical image modality. The remainder of this paper uses the terms ``anatomy'' and ``modality'', which are associated with medical image analysis, to refer to the synonymous ``content'' and ``style'' that are most common in deep learning/computer vision terminology.

\subsection{Semi-supervised segmentation} \label{subsec:semisupervised}

A powerful property of disentangled representations is that they can be applied in semi-supervised learning \citep{almahairi2018augmented}. An important application in medical image analysis is (semi-supervised) segmentation, for a recent review see \cite{cheplygina2018not}. As discussed in this review, manual segmentations are a laborious task, particularly as inter-rater variation means multiple labels are required to reach a consensus, and images labelled by multiple experts are very limited. Semi-supervised segmentation has been proposed for cardiac image analysis using an iterative approach and Conditional Random Fields (CRF) post-processing \citep{Bai2017}, and for gland segmentation using GANs \citep{ZhangYizhe2017}.

More recent medical semi-supervised image segmentation approaches include \cite{zhao2018deep} and \cite{nie2018asdnet}. \cite{zhao2018deep} address a multi-instance segmentation task in which they have bounding boxes for all instances, but pixel-level segmentation masks for only some instances. \cite{nie2018asdnet} approach semi-supervised segmentation with adversarial learning and a confidence network. Neither approaches involve learning disentangled representations of the data.

\subsection{Cardiac segmentation} \label{subsec:cardiac_segmentation}

We apply our model to the problem of cardiac segmentation, for which there is considerable literature \citep{peng2016review}. The majority of recent methods use convolutional networks with full supervision for multi-class cardiac segmentations, as seen for example in participants of workshop challenges \citep{BernardACDC}. Cascaded networks \citep{VIGNEAULT201895} are used to perform 2D segmentation by transforming the data into a canonical orientation and also by combining information from different views.
Prior information about the cardiac shape has been used to improve segmentation results \citep{OktayAnatomically}. Spatial correlation between adjacent slices has been explored \citep{Zheng3DConsistent} to consistently segment 3D volumes. Segmentation can also be treated as a regression task \citep{TAN201778}. Finally, temporal information related to the cardiac motion has been used for segmentation of all cardiac phases \citep{Qin2018_JointMotion, Bai2018_Recurrent}.

Differently from the above, in this work we focus on learning meaningful spatial  representations, and leveraging these for improved semi-supervised segmentation results, and performing auxiliary tasks.

\section{Materials and methods} \label{sec:methods}

Overall, our proposed model can be considered as an autoencoder, which takes as input a 2D volume slice $x \in X$, where $X \subset {\rm I\!R}^{H \times W \times 1}$ is the set of all images in the data, with $H$ and $W$ being the image's height and width respectively. The model generates a reconstruction through an intermediate disentangled representation. The disentangled representation is comprised of a multi-channel spatial map (a tensor) $s \in S \coloneqq \{0,1\}^{H\times W\times C}$, where $C$ is the number of channels, and a multi-dimensional continuous vector factor $z \in Z \coloneqq {\rm I\!R}^{n_z}$, where $n_z$ is the number of dimensions. These are generated respectively by two encoders, modelled as convolutional neural networks, $f_{anatomy}$ and $f_{modality}$. The two representations are combined by a decoder $g$ to reconstruct the input. In addition to the reconstruction cost, explicit supervision can be given in the form of auxiliary tasks, for example with a segmentation task using a network $h$, or with a regression task as we will demonstrate in Section \ref{subsec:lvv}. A schematic of our model can be seen in Figure \ref{fig:simpleschematic} and the detailed architectures of each network are shown in Figure \ref{fig:schematic}.

\subsection{Input decomposition}

The decomposition process yields representations for the anatomy and the modality characteristics of medical images and is achieved by two dedicated neural networks.
Whilst a decomposition could also be performed with a single neural network with two separate outputs and shared layer components, as done in  our previous work \citep{chartsias2018factorised}, we found that by using two separate networks, as also done in \cite{huang2018munit} and in \cite{lee2018diverse}, we can more easily control the information captured by each factor, and we can stabilise the behaviour of each encoder during training.

\subsubsection{Anatomical representation}

\begin{figure*}[t!]
\centering
\begin{subfigure}{\textwidth}
   \includegraphics[width=1\linewidth]{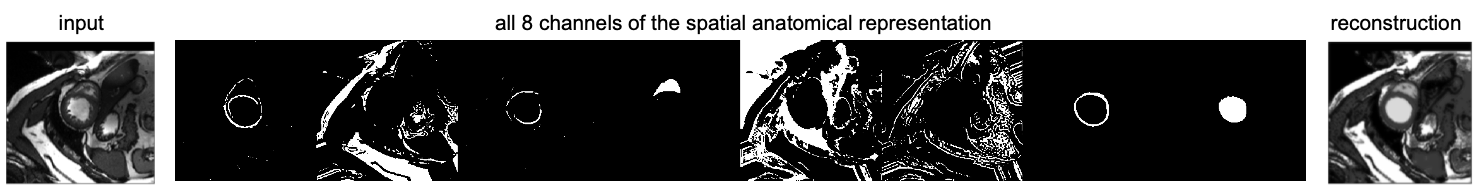}
   \caption{Anatomical representation with binary thresholding.}
   \label{fig:s} 
\end{subfigure}
\\
\begin{subfigure}{\textwidth}
   \includegraphics[width=1\linewidth]{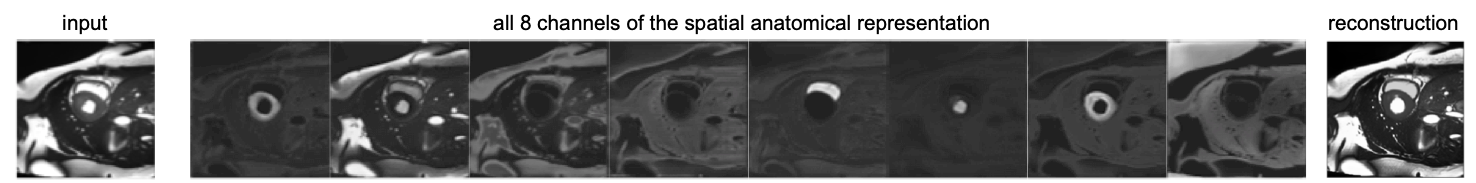}
   \caption{Anatomical representation with no binary thresholding.}
   \label{fig:norounding}
\end{subfigure}
\caption{(a) Example of a spatial representation, expressed as a multi-channel binary map. Some channels represent defined anatomical parts such as the myocardium or the left ventricle, and others the remaining anatomy required to describe the input image on the left. Observe how sparse most of the informative channels are. (b) Spatial representation with no thresholding applied. Each channel of the spatial map, also captures the intensity signal in different gray level variations and is not sparse, in contrast to Figure \ref{fig:s}. This may hinder an anatomical separation. Note that no specific channel ordering is imposed and thus the anatomical parts can appear in different order in the anatomical representations across experiments.} 
\end{figure*}

The anatomy encoder is a fully convolutional neural network that maps 2D images to spatial representations, $f_{anatomy}: X \rightarrow S$. We use a U-Net \citep{ronneberger2015u} architecture, containing downsampling and upsampling paths with skip connections between feature maps of the same size, allowing effective fusion of important local and non-local information. 

The spatial representation is a feature map consisting of a number of binary channels of the same spatial dimensions $(H, W)$ as the input image, that is $s \in \{0, 1\}^{H \times W \times C} s.t. \sum_{c=1}^{C} s_{h,w,c} = 1$ $\forall h \in \{1, \ldots, H\}, w \in \{1, \ldots, W\}$, where $C$ is the number of channels. Some channels contain individual anatomical (cardiac) sub-structures, while the other structures, necessary for reconstruction, are freely dispersed in the remaining channels.
Figure \ref{fig:s} shows an example of a spatial representation, where the representations of the myocardium, the left and the right ventricle, are clearly visible, and the remaining channels contain the surrounding image structures (albeit more mixed and not anatomically distinct).

The spatial representation is derived using a softmax activation function to force each pixel to have activations that sum to one across the channels. Since softmax functions encode continuous distributions, we binarise the anatomical representation via the operator $s \mapsto \lfloor s+0.5 \rfloor$, which acts as a threshold for the pixel values of the spatial variables in the forward pass. During back-propagation the step function is bypassed and updates are applied to the original non-binary representation, as in the straight-through operator \citep{BengioLC13}. 

Thresholding $s$ is an integral part of the model's design and offers two advantages. Firstly, it reduces the capacity of the spatial factor, encouraging it to be a representation of only the anatomy and preventing modality information from being encoded. Secondly, it enforces a factorisation of the spatial factor in distinct channels, as each pixel can only be active on one channel. To illustrate the importance of this binarisation, an example of a non-thresholded spatial factor is shown in Figure \ref{fig:norounding}. Observe, that the channels of $s$ are not sparse with variations of gray level now evident.  Image intensities are now encoded spatially, using different grayscale values, allowing a good reconstruction to be achieved without the need of a modality factor, which we explicitly want to avoid.

\subsubsection{Modality representation} \label{sec:modality_representation}

Given samples of the data $x \in X$ with their corresponding $s \in S$ (deterministically obtained by $f_{anatomy}$), we learn the posterior distribution of latent factors $z \in Z \coloneqq {\rm I\!R}^{n_z}$, $q(z|x,s)$. 

Learning this posterior distribution follows the VAE principle \citep{kingma2013auto}. In brief a VAE learns a low dimensional latent space, such that the learned latent representations match a prior distribution that is set to be an isotropic multivariate Gaussian  $p(z)=\mathcal{N}(0,1)$. A VAE consists of an encoder and a decoder. The encoder, given an input, predicts the parameters of a Gaussian distribution (with diagonal co-variance matrix). This distribution is then sampled, using the reparameterisation trick to allow learning through back propagation, and the resulting sample is fed through the decoder to reconstruct the input. VAEs are trained to minimise a reconstruction error and the KL divergence of the estimated Gaussian distribution $q(z|x,s)$ from the unit Gaussian $p(z)$, 
\begin{linenomath}\begin{equation*}
    L_{KL} = D_{KL}(q(z|x,s)\|p(z)),
\end{equation*}\end{linenomath}
where $D_{KL}(p\|q)=\int p(z)log \frac{p(z)}{q(z|x,s)}dxds$.
Once trained, sampling a vector from the unit Gaussian over the latent space and passing it through the decoder approximates sampling from the data, i.e. the decoder can be used as a generative model.

The posterior distribution is modelled with a stochastic encoder (this is analogous to the VAE encoder) as a convolutional network, which encodes the image modality, $f_{modality}: X\times S \rightarrow Z$. Specifically, the stochasticity of the encoder (for a sample $x$ and its anatomy factor $s$) is achieved as in the VAE formulation as follows: $f_{modality}(x,s)$ produces first the mean and diagonal covariance for an $n_z$ dimensional Gaussian, which is then sampled to yield the final $z$. 

\subsection{Segmentation}\label{sec:segmentation}
One important task for the model is to infer segmentation masks $m \in M \coloneqq \{0, 1\}^{H \times W \times L}$, where $L$ is the number of anatomical segmentation categories in the training dataset, out of the spatial representation. This is an integral part of the training process because it also defines the anatomical structures that will be extracted from the image. The segmentation network\footnote{Experimental results showed that having an additional segmentor network, instead of enforcing our spatial representation to contain the exact segmentation masks, improves the training stability of our method. Furthermore, it offers flexibility in that the same anatomical representation can be used for multiple tasks, such as in segmentation and the calculation of the left ventricular volume.} is a fully convolutional network consisting of two convolutional blocks followed by a final $1\times1$ convolution layer (see Figure \ref{fig:schematic}), with the goal of refining the anatomy present in the spatial maps and produce the final segmentation masks, $h: S \rightarrow M$. 

When labelled data are available, a supervised cost is employed that is based on a differentiable Dice loss \citep{Milletari2016VNetFC} between a real segmentation mask $m$ of an image sample $x$ and its predicted segmentation $h(f_{anatomy}(x))$, 
\begin{linenomath}\begin{equation*}
    L_{segm} =  1 - 2 \times \mathop{\mathbb{E}}_{x, m} \left[ \frac{\sum_{h,w,l} (m_{h,w,l} \times h(f_{anatomy}(x))_{h,w,l}) + \epsilon}{\sum_{h,w,l} (m_{h,w,l} + h(f_{anatomy}(x))_{h,w,l}) + \epsilon} \right],
\end{equation*}\end{linenomath}
where the added small constant $\epsilon$ prevents division by 0.
In a semi-supervised scenario, where there are images with no corresponding segmentations, an adversarial loss is defined using a discriminator over masks $D_M$, based on LeastSquares-GAN \citep{mao2017effectiveness}. Networks $f_{anatomy}$ and $h$ are trained to maximise the adversarial objective, against $D_M$ which is trained to minimise it,
\begin{linenomath}\begin{equation*}
    L_{adv} = \mathop{\mathbb{E}}_{x, m}\left[D_M(h(f_{anatomy}(x)))^2 + (D_M(m)-1)^2\right].
\end{equation*}\end{linenomath}
The architecture of the discriminator is based on DCGAN discriminator \citep{Radford2015UnsupervisedRL}, without Batch Normalization.

\subsection{Image reconstruction} \label{sec:image_reconstruction}

The two factors are combined by a decoder network $g$ to generate an image $y \in Y \coloneqq {\rm I\!R}^{H \times W \times 1}$ with the anatomical characteristics specified by $s$ and the imaging characteristics specified by $z$, $g: S \times Z \rightarrow Y$. The fusion of the two factors acts as an inpainting mechanism where the information stored in $z$, is used to derive the image signal intensities that will be used on the anatomical structures, stored in $s$. 

The reconstruction is achieved by a decoder convolutional network conditioned with four FiLM \citep{perez2017film} layers.
This general purpose conditioning method learns scale and offset parameters for each feature-map channel within a convolutional architecture. Thus, an affine transformation (one per channel) learned from the conditioning input is applied.

Here, a network of two fully connected layers (see Figure \ref{fig:schematic}) maps $z$ to the scale and offset values $\gamma$ and $\beta$ for each intermediate feature map $F_c$ of the decoder. Each channel of $F_c$ is modulated based on $c$ pairs $\gamma_c$ and $\beta_c$ as follows: $FiLM(F_c|\gamma_c, \beta_c) = \gamma_c \odot F_c + \beta_c$, where  element-wise multiplication ($\odot$) and addition are both broadcast over the spatial dimensions.
The decoder and FiLM parameters are learned through the reconstruction of the input images using Mean Absolute Error,
\begin{linenomath}\begin{equation*}
    L_{rec} = \mathop{\mathbb{E}}_{x}\left[\|x - g(f_{anatomy}(x), f_{modality}(x, f_{anatomy}(x)))\|_1\right].
\end{equation*}\end{linenomath}

The design of the decoding process restricts the type of information stored in $z$ to only affect the intensities of the produced image. This is important in the disentangling process as it pushes $z$ to not contain spatial anatomical information.

The decoder can also be interpreted as a conditional generative model, where different samples of $z$ conditioned on a given $s$ generate images of the same anatomical properties, but with different appearances. The reconstruction process is the opposite of the decomposition process, i.e. it learns the dependencies between the two factors in order to produce a realistic output.

\subsubsection{Modality factor reconstruction}

A common problem when training VAE is posterior collapse: a degenerate condition where the decoder is ignoring some factors. In this case, even though the reconstruction is accurate, not all data variation is captured in the underlying factors. 

In our model posterior collapse manifests when some modality information is spatially encoded within the anatomical factor.\footnote{Note that while using FiLM prevents $z$ from encoding spatial information, it does not prevent the case of posterior collapse i.e. that $s$ encodes (all or part of) the modality information.} To overcome this we use a $z$ reconstruction cost, according to which an image $y$ produced by a random $z$ sample should produce the same modality factor when (re-)encoded,
\begin{linenomath}\begin{equation*}
    L_{z_{rec}} = \mathop{\mathbb{E}}_{z, y}\left[\|z - f_{modality}(y, f_{anatomy}(y))\|_1\right].
\end{equation*}\end{linenomath}
The faithful reconstruction of the modality factor $z$ penalises the VAE for ignoring dimensions of the latent distribution and encourages each encoded image to produce a low variance Gaussian. This is in tension with the KL divergence cost which is optimal when the produced distribution is a spherical Gaussian of zero mean and unit variance. A perfect score of the KL divergence results in all samples producing the same distribution over $z$, and thus the samples are indistinguishable from each other based on $z$. Without $L_{z_{rec}}$, the overall cost function can be minimised if imaging information is encoded in $s$, thus resulting in posterior collapse. Reconstructing the modality factor prevents this, and results in an equilibrium where a good reconstruction is possible only with the use of both factors. 

\section{Experimental setup} \label{sec:experiments}
\subsection{Data}

In our experiments we use 2D images from four datasets, which have been normalised to the range [-1, 1].
\begin{enumerate}[label=(\alph*)]
    \item For the semi-supervised segmentation experiment (Section \ref{subsec:expsemisupervised}) and the latent space arithmetic (Section \ref{subsec:arithmetic}) we use data from the 2017 \textit Automatic Cardiac Diagnosis Challenge (ACDC) \citep{BernardACDC}. This dataset contains cine-MR images acquired in 1.5T and 3T MR scanners, with resolution between 1.22 and 1.68 $mm^2/pixel$ and a number of phases varying between 28 to 40 images per patient. We resample all volumes to 1.37 $mm^2/pixel$ resolution. Images are cropped to $224\times 224$ pixels. There are images of 100 patients for which manual segmentations are provided for the left ventricular cavity (LV), the myocardium (MYO) and the right ventricle (RV), corresponding to the end systolic (ES) and end diastolic (ED) cardiac phases. In total there are 1,920 images with manual segmentations (from ED and ES) and 23,530 images with no segmentations (from the remaining cardiac phases).
    
    \item We also use data acquired at Edinburgh Imaging Facility QMRI with a 3T scanner. The dataset contains cine-MR images of 26 healthy volunteers each with approximately 30 cardiac phases. The spatial resolution is 1.406 $mm^2$/pixels with a slice thickness of $6mm$, matrix size $256 \times 256$, a field of view $360mm \times 303.75 mm$, and image size $256\times 208$ pixels. This dataset is used in the semi-supervised segmentation and multi-task experiments of Sections \ref{subsec:expsemisupervised} and \ref{subsec:lvv} respectively. Manual segmentations of the left ventricular cavity (LV) and the myocardium (MYO) are provided, corresponding to the ED cardiac phase. In total there are 241 images with manual segmentations (from ED) and 8,353 images with no segmentations (from the remaining cardiac phases).
    
    \item To demonstrate multimodal segmentation and modality transformation (Section \ref{subsec:multimodal_segmentation}), as well as modality estimation  (Section \ref{subsec:multimodal_classification}), we use data from the 2017 \textit{Multi-Modal Whole Heart Segmentation} (MM-WHS) challenge, made available by \cite{zhuang2010registration}, \cite{zhuang2013challenges}, and \cite{zhuang2016multi}. This contains 40 anonymised volumes, of which 20 are cardiac CT/CT angiography (CTA) and 20 are cardiac MRI. The CT/CTA data were acquired in the axial view at Shanghai Shuguang Hospital, China, using routine cardiac CTA protocols. The in-plane resolution is about $0.78 \times 0.78 mm$ and the average slice thickness is $1.60 mm$. The MRI data were acquired at St. Thomas hospital and Royal Brompton Hospital, London, UK, using 3D balanced steady state free precession (b-SSFP) sequences, with about $2 mm$ acquisition resolution at each direction and reconstructed (resampled) into about $1 mm$.
    All data have manual segmentations of seven heart substructures:  myocardium (MYO), left atrium (LA), left ventricle (LV), right atrium (RA), right ventricle (RV), ascending aorta (AO) and pulmonary artery (PA). Data preprocessing is as in \cite{chartsias2017adversarial}.  The image size is $224\times 224$ pixels. In total there are 3,626 MR and 2,580 CT images, all with manual segmentations.
    
    \item Finally, we use cine-MR and CP-BOLD images of 10 canines to further evaluate modality estimation (Section \ref{subsec:multimodal_classification}). 2D images with an in-plane resolution of $1.25mm \times 1.25mm$ were acquired at baseline and severe ischemia (inflicted as controllable stenosis of the left-anterior descending coronary artery (LAD)) on a 1.5T Espree (Siemens Healthcare) on the same instrumented canines. The image acquisition is at short axis view, covering the mid-ventricle, and is performed using cine-MR and a flow and motion compensated CP-BOLD acquisition. The pixel resolution is $192 \times 114$ \citep{tsaftaris2013detecting}. This dataset (not publicly available) is ideal to show complex spatio-temporal effects as it images the same animal with and without disease and using two almost identical sequences with the only difference that CP-BOLD modulates pixel intensity with the level of oxygenation present in the tissue. In total there are 129 cine-MR and 264 CP-BOLD images with manual segmentations from all cardiac phases.
\end{enumerate}

\subsection{Model and training details}

The overall cost function is a composition of the individual costs of each of the model's components and is defined as:
\begin{linenomath}\begin{equation*}
L = \lambda_1 L_{KL} + \lambda_2 L_{segm} + \lambda_3 L_{adv} + \lambda_4 L_{rec} + \lambda_5 L_{z_{rec}}.
\end{equation*}\end{linenomath}

The $\lambda$ parameters are set to values: $\lambda_1$=0.01, $\lambda_2$=10, $\lambda_3$=10, $\lambda_4$=1, $\lambda_5$=1. We adopt the value of $\lambda_1$ from \cite{zhu2017toward}, that also trains a VAE for modelling intensity variability. Separating the anatomy into segmentation masks is a difficult task, and is also in tension with the reconstruction process which pushes parts with similar intensities to be in the same channels. This motivates our decision in increasing the values of the segmentation hyperparameters $\lambda_2$ and $\lambda_3$.

We set the dimension of the modality factor $n_z$=8 as in \cite{zhu2017toward} across all datasets. We also set the number of channels of the spatial factor to $C$=8 for ACDC and QMRI and increase to $C$=16 for MM-WHS, to support the increased number of segmented regions (7 in MM-WHS) and the fact that CT and MR data have different contrasts and viewpoints. This additional flexibility allows the network to use some channels of $s$ for common information across the two modalities (MR and CT) and some for unique (not common) information. 

We train using Adam \citep{KingmaB14} with a learning rate of 0.0001 and a decay of 0.0001 per epoch. 
We used a batch size of 4 and an early stopping criterion based on the segmentation cost of a validation set. All code was developed in Keras \citep{chollet2015keras}. The quantitative results of Section \ref{sec:results} are obtained through 3-fold cross validation, where each split contains a proportion of the total volumes of 70\%, 15\% and 15\% corresponding to training, validation and test sets.
SDNet implementation will be made available at \url{https://github.com/agis85/anatomy_modality_decomposition}.

\subsection{Baseline and benchmark methods} \label{subsec:baselines}

We evaluate our model's segmentation accuracy by comparing with one fully supervised and two semi-supervised methods described below:
\begin{enumerate}[label=(\alph*)]
    \item We use \textbf{U-Net} \citep{ronneberger2015u} as a fully supervised baseline because of its effectiveness in various medical segmentation problems, and also since it is frequently used by the participants of the two cardiac challenges MM-WHS and ACDC. It's architecture follows the one proposed in the original paper, and is the same as the SDNet's anatomy encoder for fair comparison. 
    \item We add an adversarial cost using a mask discriminator to the fully-supervised U-Net, enabling its use in semi-supervision. This can also be considered as a variant of SDNet without the reconstruction cost. We refer to this method as \textbf{GAN} in Section \ref{sec:results}.
    \item We also use the \textbf{self-train} method of \cite{Bai2017}, which proposes an iterative method of using unlabelled data to retrain a segmentation network. In the original paper a Conditional Random Field (CRF) post-processing is applied. Here, we use U-Net as a segmentation network (such that the same architecture is used by all baselines) and we do not perform any post-processing for a fair comparison with the other methods we present.
\end{enumerate}

To permit comparisons, training of the baselines uses the same hyperparameters, such as learning rate decay, optimiser, batch size, and early stopping criterion, as used for SDNet.

\section{Results and discussion} \label{sec:results}

We here present and discuss quantitative and qualitative results of our method in various experimental scenarios. Initially, multi-class semi-supervised segmentation is evaluated in Section \ref{subsec:expsemisupervised}. Subsequently, Section \ref{subsec:lvv} demonstrates multi-task learning with the addition of a regression task in the training objectives. In Section \ref{subsec:multimodal_segmentation}, SDNet is evaluated in a multimodal scenario by concurrently segmenting MR and CT data.
In Section \ref{subsec:multimodal_classification} we investigate whether the modality factor $z$ captures multimodal information. Finally, Section \ref{subsec:arithmetic} demonstrates properties of the factorisation using latent space arithmetic, in order to show how $z$ and $s$ interact to reconstruct images.

\subsection{Semi-supervised segmentation} \label{subsec:expsemisupervised}

We evaluate the utility of our method in a semi-supervised experiment, in which we combine labelled images with a pool of unlabelled images to achieve multi-class semi-supervised segmentation. Specifically, we explore the sensitivity of SDNet and the baselines of Section \ref{subsec:baselines} to the number of labelled examples, by training with various numbers of labelled images. Our objective is to show that we can achieve comparable results to a fully supervised network using fewer annotations.

To simulate a more realistic clinical scenario, sampling of the labelled images does not happen over the full image pool, but at a subject level: initially, a number of subjects are sampled, and then all images of these subjects constitute the labelled dataset.
The number of unlabelled images is fixed and set equal to 1200 images: these are sampled at random from all subjects of the training set and from cardiac phases other than End Systole (ES) and End Diastole (ED) (for which no ground truth masks exist). The real segmentation masks used to train the mask discriminator are taken from the set of image-mask pairs from the same dataset.

In order to test the generalisability of all methods to different types of images, we use two cine-MR datasets: ACDC which contains masks of the LV, MYO and RV; and QMRI which contains masks of the LV and MYO. Spatial augmentations by rotating inputs up to \ang{90} are applied to experiments using ACDC data to better simulate the orientation variability of the dataset. No augmentations are applied in experiments using QMRI data since all images maintain a canonical orientation. No further augmentations have been performed to fairly compare the effect of the different methods.

\begin{figure*}[h!]
\centering
\includegraphics[width=\linewidth]{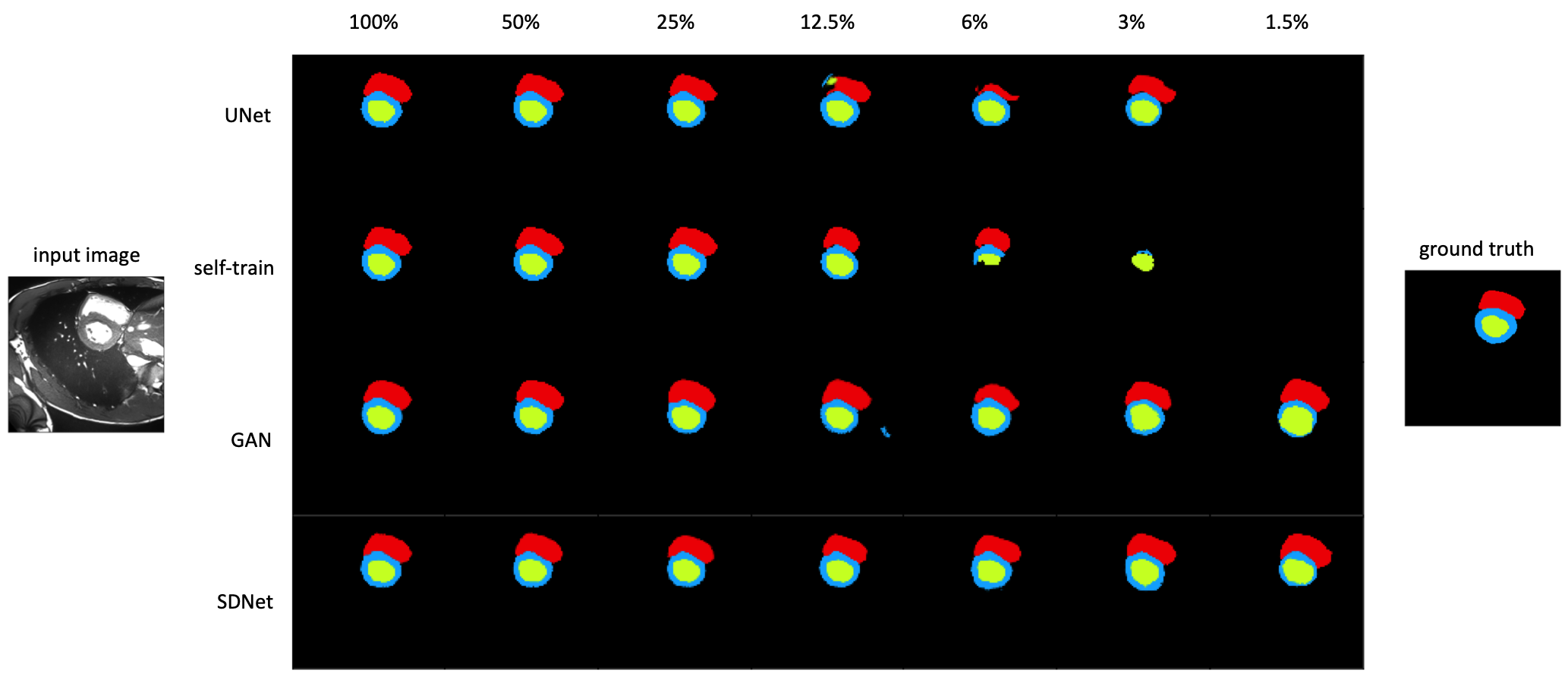}
\caption{Segmentation example for different numbers of labelled images from the ACDC dataset. Blue, green and red show the models prediction for MYO, LV and RV respectively.}
\label{fig:segmentation_acdc}
\end{figure*}

We present the average cross-validation Dice score (on held out test sets) across all labels, as well as the Dice score for each label separately, and the corresponding standard deviations. Note that images from a given subject can only be present in exactly one of the training, validation or test sets. Table \ref{table:semisupervised_acdc} contains the ACDC results for all labels, MYO, LV and RV respectively, and Table \ref{table:semisupervised_qmri} contains the QMRI results for all labels, MYO, and LV respectively. The test set for each fold contains 280 images of ED and ES phases, belonging to 15 subjects for ACDC, and 35 images of the ED phase belonging to 4 subjects for QMRI.
The best results are shown in bold font, and an asterisk indicates statistical significance at the $5\%$ level, compared to the second best result, computed using a paired t-test. In both tables the lowest amount of labelled data ($1.5\%$ for Table \ref{table:semisupervised_acdc} and $6\%$ for Table \ref{table:semisupervised_qmri}) correspond to images selected from one subject. Segmentation examples for ACDC data using different number of labelled images are shown in Figure \ref{fig:segmentation_acdc}, where different colours are used for the different segmentation classes.

For both datasets, when the number of annotated images is high, then all methods perform equally well, although our method achieves the lowest variance. In Table \ref{table:semisupervised_acdc} the performance of the supervised (U-Net) and self-trained methods decreases when the number of annotated images reduces below $12.5\%$, since the limited annotations are not sufficiently representative of the data. When using data from one or two subjects, these two methods which mostly rely on supervision fail with a Dice score below 55\%. On the other hand, even when the number of labelled images is small, adversarial training used by SDNet and GAN helps maintaining a good performance. The reconstruction cost used by our method further regularises training and consistently produces more accurate results, with Dice scores equal to 73\%, 77\% and 78\% for 1.5\%, 3\% and 6\% labels respectively, that are also significantly better, with p-values 0.0006, 0.02, and 0.002, in a paired t-test. 

It is interesting to compare the performance of SDNet with our previous work \citep{chartsias2018factorised}. We therefore modify our previous model for multi-class segmentation and repeat the experiment for the ACDC dataset. We compute the Dice scores and standard deviations for 100\%, 50\%, 25\%, 12.5\%, 6\%, 3\%, and 1.5\% of labelled data to be respectively $79\pm7\%$, $75\pm8\%$, $79\pm7\%$, $77\pm10\%$, $75\pm9\%$, $66\pm15\%$, and $59\pm13\%$. Comparing with the results of Table \ref{table:semisupervised_acdc}, SDNet significantly outperforms our previous model (at the 5\% level, paired t-test).

On the smaller QMRI dataset, the segmentation results are seen in Table \ref{table:semisupervised_qmri}, and correspond to two masks instead of three. When using annotated images from just a single subject (corresponding to 6\% of the data the lowest possible), the performance of the supervised method reduces by almost 50\% compared to when using the full dataset. SDNet and GAN both maintain a good performance of 75\% and 79\%, with no significant differences between them.

\begin{table*}[!t]
\begin{center}
\caption{Dice score (\%) on ACDC for MYO, LV, RV, and average. Standard deviations are shown as subscripts. The models are trained with 1200 unlabelled and different fraction of labelled images (each one corresponding to a proportion of selected subjects). For each of the three components and the average separately, the best result is shown in bold font and an asterisk indicates statistical significance at the 5\% level compared to the second best method in the same row/component.} 
\label{table:semisupervised_acdc}
\setlength\tabcolsep{4.5pt}
\begin{tabular}{r|cccc|cccc|cccc|cccc}
\toprule
labels & \multicolumn{4}{c|}{\textbf{U-Net}} & \multicolumn{4}{c|}{\textbf{GAN}} & \multicolumn{4}{c|}{\textbf{self-train}} & \multicolumn{4}{c}{\textbf{SDNet}} \\
& MYO & LV & RV & avg & MYO & LV & RV & avg & MYO & LV & RV & avg & MYO & LV & RV & avg \\
\midrule
100\%  & $83_7$    & $88_6$    & $79_{10}$ & \boldmath{$85_7$}    & $82_6$    & $87_6$    & $75_8$    & $83_5$    & \boldmath{$84_7$}    & \boldmath{$89_5$}    & \boldmath{$82_8$}    & \boldmath{$85_6$}    & \boldmath{$84_5$}    & $88_4$ & $78_8$ & $84_5$ \\
50\%   & \boldmath{$83_7$}    & \boldmath{$87_7$}    & \boldmath{$79_{10}$} & \boldmath{$85_7$}    & $81_7$    & $86_6$    & $75_{10}$ & $82_7$    & $80_{10}$ & $85_{10}$ & $78_{11}$ & $82_8$    & \boldmath{$83_6$}    & \boldmath{$87_7$} & $77_9$ & $83_6$ \\
25\%   & $77_9$    & $82_9$    & $67_{14}$ & $75_{11}$ & $78_9$    & \boldmath{$85_8$}    & $72_{11}$ & $79_8$    & $76_{13}$ & \boldmath{$85_{10}$} & $70_{15}$ & $78_{11}$ & \boldmath{$80_7^*$}    & \boldmath{$85_6$}    & \boldmath{$73_{11}$} & \boldmath{$81_6^*$} \\
12.5\% & $71_{13}$ & $80_{13}$ & $61_{17}$ & $70_{13}$ & $78_8$    & \boldmath{$85_6$}    & \boldmath{$69_{13}$} & $79_8$    & $63_{17}$ & $77_{13}$ & $57_{21}$ & $67_{15}$ & \boldmath{$79_8$}    & \boldmath{$85_7$}    & \boldmath{$69_{13}$} & \boldmath{$80_8$} \\
6\%    & $63_{12}$ & $76_{13}$ & $56_{22}$ & $65_{13}$ & $75_{11}$ & $81_{11}$ & $69_{13}$ & $75_{12}$ & $46_{27}$ & $59_{23}$ & $34_{18}$ & $47_{23}$ & \boldmath{$77_9$}    & \boldmath{$83_{10}$} & \boldmath{$71_{12}$} & \boldmath{$78_9^*$} \\
3\%    & $55_{19}$ & $66_{20}$ & $46_{20}$ & $52_{18}$ & $73_{32}$ & $79_{10}$ & $67_{14}$ & $75_{10}$ & $20_{15}$ & $35_{20}$ & $22_{14}$ & $24_{15}$ & \boldmath{$76_7^*$}    & \boldmath{$82_8^*$}    & \boldmath{$68_{14}$} & \boldmath{$77_8^*$} \\
1.5\%  & $26_{19}$ & $33_{21}$  & $35_{17}$ & $21_{19}$  & $67_{21}$ & $78_{11}$  & $63_{12}$ & $67_{12}$  & $11_{10}$ & $19_{14}$  & $25_{12}$ & $16_{11}$ & \boldmath{$70_{12}$}  & \boldmath{$77_{13}$} & \boldmath{$64_{15}$}  & \boldmath{$73_{12}^*$} \\
\bottomrule
\end{tabular}
\end{center}
\end{table*}

\begin{table*}[!t]
\begin{center}
\caption{Dice score (\%) on QMRI for MYO, LV, and average. Standard deviations are shown as subscripts. The models are trained with 1200 unlabelled and different fraction of labelled images (each one corresponding to a proportion of selected subjects). For each of the two components and the average separately, the best result is shown in bold font and an asterisk indicates statistical significance at the 5\% level compared to the second best method in the same row/component.}
\label{table:semisupervised_qmri}
\centering
\begin{tabular}{r|ccc|ccc|ccc|ccc}
\toprule
labels & \multicolumn{3}{c|}{\textbf{U-Net}} & \multicolumn{3}{c|}{\textbf{GAN}} & \multicolumn{3}{c|}{\textbf{self-train}} & \multicolumn{3}{c}{\textbf{SDNet}} \\
& MYO & LV & avg & MYO & LV & avg & MYO & LV & avg & MYO & LV & avg \\
\midrule
100\%  & $72_9$    & $90_6$    & $83_7$    & \boldmath{$75_7$} & \boldmath{$93_3$} & \boldmath{$86_4$} & \boldmath{$75_9$} & $92_5$ & \boldmath{$86_7$} & \boldmath{$75_6$} & \boldmath{$93_4$} & \boldmath{$86_4$} \\
50\%   & $72_{15}$ & $82_{18}$ & $74_{15}$ & $71_9$ & $86_7$ & $83_5$ & $62_{11}$ & $88_9$ & $79_9$ & \boldmath{$73_6$} & \boldmath{$90_5$} & \boldmath{$84_5$} \\
25\%   & $54_{14}$ & $80_9$    & $69_{10}$ & \boldmath{$68_7$} & $86_7$ & \boldmath{$81_5$} & $36_{22}$ & $56_{29}$ & $49_{26}$ & $66_7$ & \boldmath{$88_7$} & $80_8$ \\
12.5\% & $52_{11}$ & $81_6$    & $65_7$    & $68_8$ & $88_6$ & $79_7$ & $42_{16}$ & $64_{14}$ & $58_{14}$ & \boldmath{$67_9$} & \boldmath{$88_6$} & \boldmath{$80_7$} \\ 
6\%    & $21_{14}$ & $43_{28}$ & $43_{20}$ & $64_9$ & $84_10$ & $75_{10}$ & $8_{6}$ & $21_{11}$ & $13_7$ & \boldmath{$65_7$} & \boldmath{$87_{10}$} & \boldmath{$79_5$} \\
\bottomrule
\end{tabular}
\end{center}
\end{table*}

\subsection{Left ventricular volume} \label{subsec:lvv}

It is common for clinicians to not manually annotate all endocardium and epicardium contours for all patients if it is not necessary. Rather, a mixture of annotations and other metrics of interest will be saved at the end of the study in the electronic health record. For example, we can have a scenario with images of some patients that contain myocardium segmentations and some images with the value of their left ventricular volume. Here we test our model in such a multi-task scenario and show that we can benefit from such auxiliary and mixed annotations. We will evaluate, firstly whether our model is capable of predicting a secondary output related to the anatomy (the volume of the left ventricle), and secondly whether this secondary task improves the performance of the main segmentation task.

Using the QMRI dataset, we first calculate the ground truth left ventricular volume (LVV) for each patient as follows: for each 2D slice, we first sum the pixels of the left ventricular cavity, then multiply this sum with the pixel resolution to get the corresponding area and then multiply the result with the slice thickness to get the volume occupied by each slice. The final volume is the sum of all individual slice volumes. 

Predicting the LVV as another output of SDNet follows a similar process to the one used to calculate the ground truth values. We design a small neural network consisting of two convolutional layers (each having a $3\times3\times16$ kernel followed by a ReLU activation), and two fully connected layers of 16 and 1 neurons respectively, both followed by a ReLU activation. This network regresses the sum of the pixels of the left ventricle, taking as input the spatial representation. The predicted sum can then be used to calculate the LVV offline.

Using a pre-trained model of labelled images corresponding to one subject (last row in Table \ref{table:semisupervised_qmri} with 6\% labels), we fine-tune the whole model whilst training the area regressor using ground truth values from 17 subjects. We find the average LVV over the test volumes equal to 138.57$mL$ (standard deviation of 8.8), and the ground truth LVV equal to 139.23$mL$ (standard deviation of 2.26), with no statistical difference between them in a paired t-test. Both measurements agree with the normal LVV values for ED cardiac phases, which was reported as 143$mL$ in a large population study \citep{Bai2018}. 
The multi-task objective used to fine-tune the whole model also benefits test segmentation accuracy, which is raised from 75.6\% to 83.2\% (statistically significant at the 5\% level).~\footnote{The multi-task objective in fact benefits the Dice score (statistically significant at the 5\% level)} for both labels individually: MYO accuracy rises from 63.3\% to 70.6\% and LV accuracy rises from 81.9\% to 89.9\%. 
While this is for a single split, observe that using LVV as an auxiliary task effectively brought us closer to the range of having 50\% annotated masks (second row in Table \ref{table:semisupervised_qmri}). Thus, auxiliary tasks, such as LVV prediction, which is related to the endocardial border segmentation, can be used to train models in a multi-task setting and leverage supervision present in typical clinical settings.

\subsection{Multimodal learning} \label{subsec:multimodal_segmentation}

By design, our model separates the anatomical factor from the image modality factor. As a result, it can be trained using multimodal data, with the spatial factor capturing the common anatomical information and the non-spatial factor capturing the intensity information unique to each image's particular modality. Here we evaluate our model using a multimodal MR and CT input to achieve segmentation (Section \ref{subsubsec:multimodal_segmentation}) and modality transformation (Section \ref{subsubsec:multimodal_synthesis}).

Both these tasks rely on learning consistent anatomical representations across the two modalities. However, it is well known that MR and CT have different contrasts that accentuate different tissue properties and may also have different views. Thus, we would expect some channels of the anatomy factor to be used in CT but not in MRI whereas some to be used by both. This disentanglement of information captures both differences in tissue contrasts but also differences in view when parts of the anatomy are not visible in all slice positions of a 3D volume. 

\begin{figure*}[t]
\centering
\includegraphics[width=\linewidth]{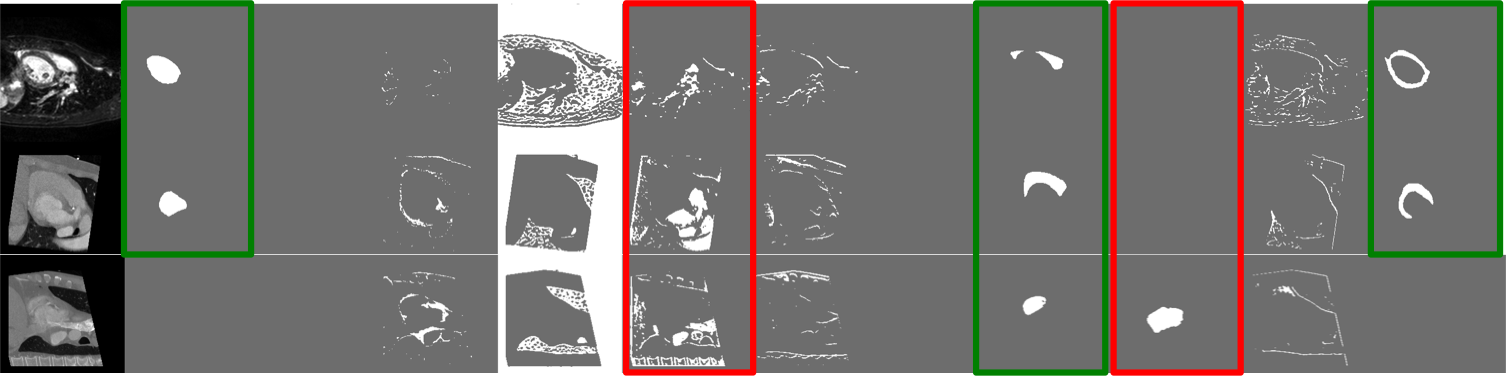}
\caption{
Example of anatomical representations from one MR and two CT images respectively. Green boxes mark common spatial information captured in the same channels, whereas red boxes mark information present in one but not the other modalities.
}
\label{fig:mrct_lr}
\end{figure*}

This is illustrated in Figure \ref{fig:mrct_lr}, which shows three example anatomical representations from one MR and two CT images, and specifically marks common anatomical factors that are captured in the same respective channels, and unique factors that are captured in different channels.

\subsubsection{Multimodal segmentation} \label{subsubsec:multimodal_segmentation}
We train SDNet using MR and CT data with the aim to improve learning of the anatomical factor from both MR and CT segmentation masks. In fact, we show below that when mixing data from MR and CT images, we improve segmentation compared to when using each modality separately. Since the aim is to specifically evaluate the effect of multimodal training in segmentation accuracy, unlabelled images are not considered here as part of the training process, and the models are trained with full supervision only. 

In Table \ref{table:multimodal} we present the Dice score over held out MR and CT test sets, obtained when training a model with differing amounts of MR and CT data. Results for 12.5\% of data correspond to images obtained from one subject. Training with both data leads to improvements in both individual MR and CT performances. This is the case even when we add 12.5\% of CT on 100\% of MR, and vice versa; this improves MR performance (from 75\% to 76\%, not statistically significant, although improvement becomes significant as more CT are added), but also CT performance (from 77\% to 81\%, statistically significant).

We also train using different mixtures of MR and CT data, but keeping the total amount of training data fixed. In the CT case, we observe that Dice ranges between 77\% (at 100\%) and 65\% (at 12.5\%). This shows that CT segmentation clearly benefits from training alongside MR, since when training on CT alone with 12.5\%, the corresponding Dice is 23\%. In the MR case, we observe that Dice ranges between 75\% (at 100\%) and 49\% (at 12.5\%). Here, the relative reduction is larger than in the CT case, however MR training at 12.5\% also benefits from the CT data, since the Dice when training on 12.5\% MR alone is 27\%. Furthermore, the Dice score for the other proportions of the data is relatively stable with a range of 69\% to 74\% for CT, and a range of 67\% to 75\% for MR. 

In both experimental setups, whether the total number of training data is fixed or not, having additional data even when coming from another modality helps. This can have implications for current or new datasets of a rare modality, which can be augmented with data from a more common modality.

\begin{table}[!t]
\begin{center}
\centering
\caption{Dice score (\%) on MM-WHS (LV, RV, MYO, LA, RA, PA, AO) data, when training with different mixtures of MR and CT data. Standard deviations are shown as subscripts.}
\begin{tabular}{cc|cc}
\toprule
\textbf{MR train} & \textbf{CT train} & \textbf{MR test} & \textbf{CT test} \\
\midrule
100\%  & 100\%  & $78_{5}$ & $80_{1}$ \\
100\%  & 12.5\% & $76_{3}$ & $56_{6}$ \\
12.5\% & 100\%  & $39_{7}$ & $81_{1}$ \\
12.5\% & 0\%    & $27_{12}$ & -     \\
0\%    & 12.5\% & -     & $23_{7}$ \\
100\%  & 0\%    & $75_{3}$ & -   \\
87.5\% & 12.5\% & $74_{5}$ & $65_{6}$ \\
75\%   & 25\%   & $75_2$ & $69_3$ \\
62.5\% & 37.5\% & $72_2$ & $69_2$ \\
50\%   & 50\%   & $68_5$ & $73_3$ \\
37.5\% & 62.5\% & $67_4$ & $73_4$ \\
25\%   & 75\%   & $67_6$ & $74_3$ \\
12.5\% & 87.5\% & $49_7$ & $73_6$ \\
0\%    & 100\%  & - & $77_{4}$ \\
\bottomrule
\end{tabular}
\label{table:multimodal}
\end{center}
\end{table}

\subsubsection{Modality transformation} \label{subsubsec:multimodal_synthesis}

Although our method is not specifically designed for modality transformations, when trained with multimodal data as input, we explore cross-modal transformations by mixing the disentangled factors. This mixing of factors is a special case of latent space arithmetic that we demonstrate concretely in Section \ref{subsec:arithmetic}.
 We combine different values of the modality factor with the same fixed anatomy factor to achieve representations of the anatomy corresponding to two different modalities. 

\begin{figure}[b!]
\centering
\includegraphics[width=\linewidth]{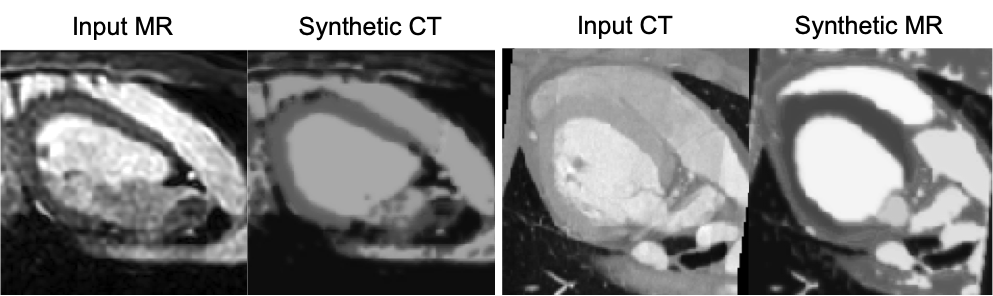}
\caption{Modality transformation between MR and CT when a fixed anatomy is combined with a modality vector derived from each imaging modality. Specifically let $x_{mr}, x_{ct}$ be $MR$ and $CT$ images respectively. The left panel of the figure shows the original MR image $x_{mr}$, and a `reconstruction' of $x_{mr}$ using the modality component derived from $x_{ct}$, i.e. $g(f_{anatomy}(x_{mr}), f_{modality}(x_{ct}, f_{anatomy}(x_{ct})))$. The right panel of the figure shows the original CT image $x_{ct}$, and a `reconstruction' of $x_{ct}$ using the modality component derived from $x_{mr}$, i.e. $g(f_{anatomy}(x_{ct}), f_{modality}(x_{mr}, f_{anatomy}(x_{mr})))$.}
\label{fig:mrct_transfer}
\end{figure}

To illustrate this we use the model trained with 100\% of the MR and CT in the MM-WHS dataset and demonstrate transformations between the two modalities. In Figure \ref{fig:mrct_transfer} we synthesise CT images from MR (and MR from CT) by fusing a CT modality vector $z$ with an anatomy $s$ from an MR image (and vice versa).  We can readily see how the transformed images capture intensity characteristics typical of the domain.

\subsection{Modality type estimation} \label{subsec:multimodal_classification}

Our premise is that the learned modality factor $z$ captures imaging specific information.  We assess this in two different settings using multimodal MR and CT data and also cine-MR and CP-BOLD MR data.

Taking one of the trained models of Table \ref{table:multimodal} corresponding to a split with 100\% MR (14 subjects of 2,837 images) and 100\% CT images (14 subjects of 1,837 images)\footnote{The results are based on a single split for ease of interpretation as between different splits we cannot relate the different z dimensions.}, we learn posthoc a logistic regression classifier (using the same training data) to predict the image modality (MR or CT) from the modality factor $z$. The learned regressor is able to correctly classify the input images as CT or MR, on a held out test set (3 subjects of 420 images for MR and 3 subjects of 387 images for CT) 92\% of the time. To find whether there is a single $z$ dimension that captures best this binary semantic component (MR or CT) we repeat 8 independent experiments training 8 single input logistic regressors, one for each dimension of $z$. We find that $z_5$ obtains an accuracy of 82\%, whereas the remaining dimensions vary from 42\% to 66\% accuracy. Thus, a single dimension (in this case $z_5$) captures most of the intensity differences between MR and CT which are global and affect all areas of the image.

In a second complementary experiment we perform the same logistic regression classification to discriminate between cine-MR and CP-BOLD MR images (which are also cine, but contain additionally oxygen-level dependent contrast). Here, SDNet and the logistic regression model are trained using 95 cine-MR and 214 CP-BOLD images from 7 subjects, and evaluated on a test set of 27 and 31 images from 1 subject respectively. Unlike MR and CT which are easy to differentiate due to differences in signal intensities across the whole anatomy, BOLD and cine exhibit subtle spatially and temporally localised differences that are modulated by the amount of oxygenated blood present (the BOLD effect) and the cardiac cycle and these are most acute in the heart.\footnote{These subtle spatio-temporal differences can detect myocardial ischemia at rest as demonstrated in \cite{bevilacqua2016dictionary, tsaftaris2013detecting}.}
Even here the classifier can detect BOLD presence with 96\% accuracy, when all dimensions of $z$ are used. When each $z$ dimension is used separately, accuracy ranges between 47\% and 65\%, and thus no single $z$ dimension globally captures the presence (or lack) of BOLD contrast.

These findings are revealing and have considerable implications.  First they show that our modality factor $z$ does capture modality specific information which is obtained completely unsupervised, and depending on context and complexity of the imaging modality, a single $z$ dimension may capture it almost completely (in the case of MR/CT). 
This also implicitly suggests that spatial information may be captured only in $s$.\footnote{It is possible to detect the modality from the anatomical factor alone. If there are systematic differences between the modalities, this can be exploited by a classifier for detection. However, in this case the modality information is not actually \emph{contained} in the anatomy factor.}

More importantly, it opens the question of how the spatial and modality factors interact to reproduce the output.  We address these questions below using latent space arithmetic.

\subsection{Latent space arithmetic} \label{subsec:arithmetic}
Herein we demonstrate the properties of the disentanglement by separately examining the effects of anatomical and modality factors on the synthetic images and how modifications of each alter the output.  
For these experiments we consider the model from Table \ref{table:semisupervised_acdc}, trained on ACDC using 100\% of the labelled training images.

\begin{figure}[t!]
\centering
\includegraphics[width=\linewidth]{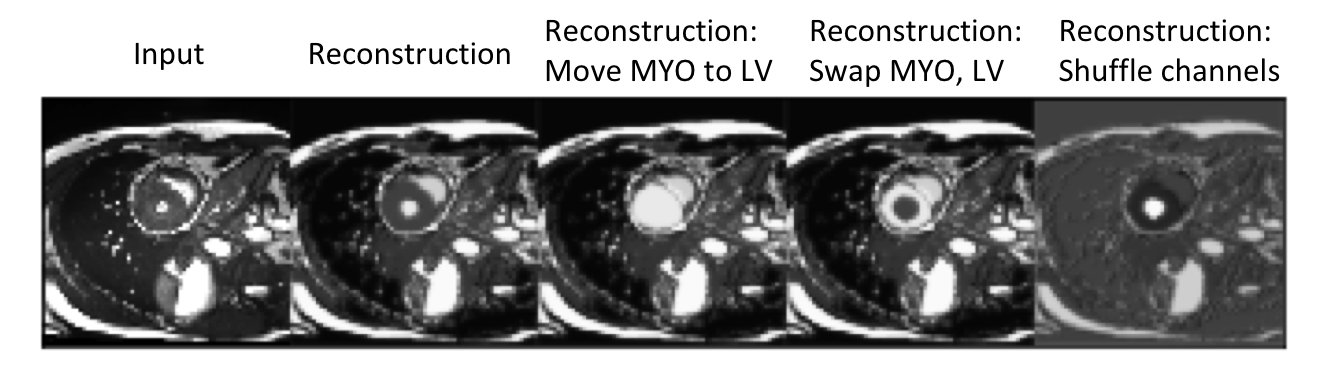}
\caption{Reconstructions of an input image, when re-arranging the channels of the spatial representation. The images from left to right are: the input, the original reconstruction, the reconstruction when moving the MYO to the LV channel, the reconstruction when exchanging the content of the MYO and the LV channels, and finally a reconstruction obtained after a random permutation of the channels.}
\label{fig:reconstructions1}
\end{figure}

\noindent \textit{Arithmetic on the spatial factor $s$:} We start with the spatial factor and in Figure \ref{fig:reconstructions1} we alter the content of the spatial channels to qualitatively see how the decoder has learned an association between the position of each channel and different signal intensities of the anatomical parts. In all these experiments the $z$ factor remains the same. The first two images show the input and the original reconstruction. The third image is produced by adding the MYO spatial channel with the LV spatial channel and by nulling (zeroing) the MYO channel. We can see that the intensity of the myocardium is now the same as the intensity of the left ventricle. In the fourth image, we swap the channels of the MYO with the one of the LV, resulting in reverse intensities for the two substructures. Finally, the fifth image is produced by randomly shuffling the spatial channels. 

\noindent \textit{Arithmetic on the modality factor $z$:} Next, we examine the information captured in each dimension of the modality factor. Since the modality factor follows a Gaussian distribution, we can draw random samples or interpolate between samples in order to generate new images. In this analysis, an image $x$ is firstly encoded to factors $s$ and $z$. Since the prior over $z$ is an 8-dimensional unit Normal distribution, $99.7\%$ of its probability mass lies within three standard deviations of the mean. As a result, the probability space is almost fully covered by values in the range $[-3, 3]$. By interpolating each $z$-dimension between $-3$ and $3$, and whilst keeping the values of the remaining dimensions and $s$ fixed, we can decode synthetic images that will show the variability induced by every $z$-dimension.

\begin{figure}[t!]
\centering
\includegraphics[width=\linewidth]{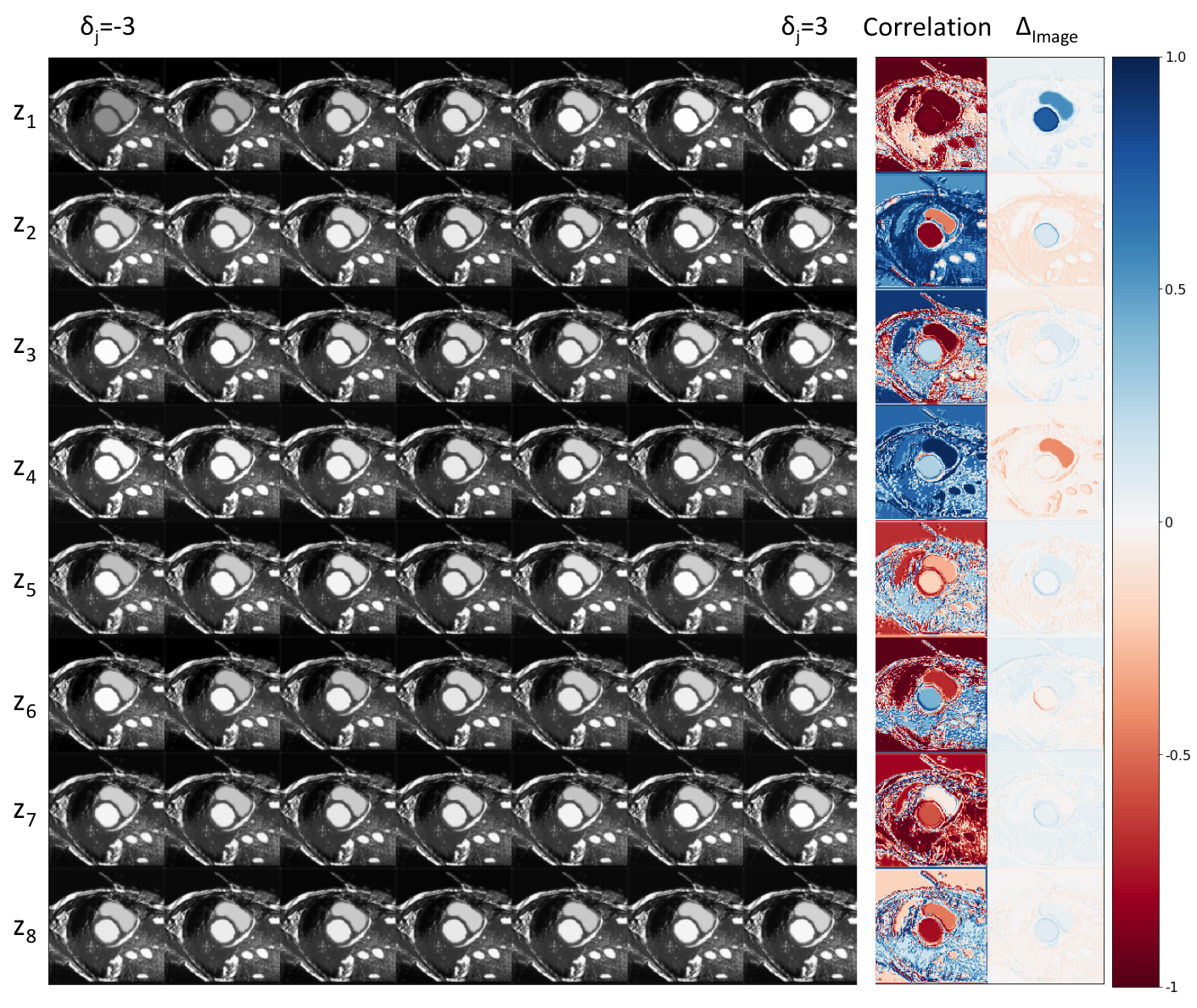}
\caption{Reconstructions when interpolating between $z$ vectors. Each row corresponds to images obtained by changing the values of a single $z$-dimension. The final two columns (correlation and $\Delta_{image}$) indicate areas of the image mostly affected by this change in $z$.}
\label{fig:reconstructions2}
\end{figure}

To achieve this we consider a grid where each $z$ dimension is considered over 7 fixed steps from $-3$ and $3$.  Each row of the grid corresponds to one of the 8 $z$ dimensions, whereas a column a specific $z$-th value in the range $[-3,3]$. This grid is visualised in Figure \ref{fig:reconstructions2}. 

Mathematically described, for $i\in \{1, 2, \ldots, 8\}$ and $j\in \{1, 2, \ldots, 7\}$, an image in the $i^{th}$ row and $j^{th}$ column of the grid is $g(s, z \odot v_{i} + (1-v_{i}) \odot \delta_{j})$, where $\odot$ denotes element-wise multiplication, $v_{i}$ is a vector of length 8 with all entries 1 except for a 0 in the $i^{th}$ position, and $\delta_{j} = -3+6(j-1)$.

In order to assess the effect of $z_i$ (the $i^{th}$ dimension of $z$) on the intensities of the synthetic results, we calculate a correlation image and a difference image (for every row of results). The value of each pixel in the correlation image is calculated using the Pearson correlation coefficient between the interpolation values of a $z_i$ and the intensity values of the synthetic images for this pixel. 

\[
    \rho_{{z_i}, y_{h,w}} = \frac{\sum_{j=1}^7(z_i^j - \bar{z}_i)(y_{h,w}^j - \bar{y}_{h,w})}{\sigma_{z_i} \sigma_{y_{h,w}}} \; \forall \; h, w \in H, W,
\]
where $h,w$ are the height and width position of a pixel, $\bar{z}_i$ is the mean value of $z_i$, $\bar{y}_{h,w}$ is the mean value of a pixel across the interpolated images. The difference image is calculated for each row by subtracting the image in the last column position on the grid ($\delta_j=3$) with the first position on the grid ($\delta_j=-3$).~\footnote{Note that in order to keep the correlation and the difference image in the same scale [-1, 1], we rescale the images from [-1, 1] to the [0,1], which does not have any effect on the results.}

In Figure \ref{fig:reconstructions2}, the correlation images show large positive or negative correlation between each $z$ dimension and most pixels of the input image, demonstrating that $z$ mostly captures global image characteristics. However, local correlations are also evident for example between $z_1$ and all pixels of the heart, between $z_4$ and the right ventricle and between $z_5$ and the myocardium. However, different magnitude changes are evident, as the difference image in the last last column of Figure \ref{fig:reconstructions2} shows. $z_1$ and $z_4$ seem to alter significantly the local contrast.

\subsection{Factor sizes} \label{subsubsec:multimodal_factor}

While throughout the paper we used $C=8$ and $n_z=8$, it is worthwhile discussing the effects of these important hyperparameters as they determine the capacity of the model. 

We have found through experiments that when $C>8$ many channels are all zero. This additional capacity is helpful when we use multimodal data, as for example in the MR/CT experiments, where $C=16$. This allows to capture information common and unique across the two modalities in different $s$-channels see Figure~\ref{fig:mrct_lr}). On the other hand, making $C$ small ($C<4$) we find that the model does not have enough capacity (for example an SDNet with $C=4$ trained at 100\% labels has Dice performance $68.1\pm8\%$, a drop compared to 84\% when $C=8$, that is also statistically significant at 5\%).

We used $n_z=8$ inspired by related literature \citep{zhu2017toward}.
Experiments with similar values of $n_z$ maintain the segmentation performance, though this is decreased for high values of $n_z$. Specifically, an SDNet with 4, 32, and 128 dimensions trained at 100\% labels has Dice $84\pm5\%$, $83\pm6\%$, and $82\pm6\%$, respectively. Compared to 84\% when $n_z=8$, the results for $n_z=4$ and $n_z=32$ are similar, but the result for $n_z=128$ is worse (and also statistically significant at 5\%), suggesting that the additional dimensions may negatively affect training and do not store extra information.
To assess this we used the methodology in \cite{burgess2018understanding} to find the capacity of each $z$-dimension, which is also a measure of informativeness. This is calculated using the average variance per dimension, where a smaller variance indicates higher capacity. A variance near 1 (with a mean=0) would indicate that this dimension encodes a Normal distribution for any datapoint, and thus, according to \cite{burgess2018understanding}, is uninformative and points to encoding the average of the distribution mode. Using this analysis, for $n_z=128$ we observed that two $z$-dimensions each had variance of 0.88, while the remaining 126 had an average variance of 0.91. Repeating this analysis for $n_z=32$, $n_z=8$ and $n_z=4$ we get the following results. For $n_z=32$, two dimensions each has variances 0.78 and 0.79, while the remaining 30 dimensions have an average variance of 0.81. For $n_z=8$, two $z$-dimensions each has variances 0.63 and 0.73, while the remaining 6 have an average variance of 0.75. Finally for $n_z=4$, two dimensions have variances 0.62 and 0.65, and the average variance of the other two is 0.77, which are similar to the results of $n_z=8$. This analysis shows that with smaller $n_z$, more informative content is captured in the individual $z$-dimensions, and thus a high $n_z$ is redundant for this particular task.


\section{Conclusion} \label{sec:conclusion}

We have presented a method for disentangling medical images into a spatial and a non-spatial latent factor, where we enforced a semantically meaningful spatial factor of the anatomy and a non-spatial factor encoding the modality information. To the best of our knowledge, maintaining semantics in the spatial factor has not been previously investigated. Moreover, through the incorporation of a variational autoencoder, we can treat our method as a generative model, which allows us to also efficiently model the intensity variability of medical data. 

We demonstrated the utility of our methodology in a semi-supervised segmentation task, where we achieve high accuracy even when the amount of labelled images is substantially reduced. We also demonstrated that the semantics of our spatial representation mean it is suitable for secondary anatomically-based tasks, such as quantifying the left ventricular volume, which not only can be accurately predicted, but also improve the accuracy of the primary task in a multi-task training scenario. 
We also show that the factorisation of the model presented can be used in multimodal learning, where both anatomical and imaging information can be encoded to create synthetic MR and CT images, using even small fractions of CT and MR input images, respectively.

The broader significance of our work is the disentanglement of medical image data into meaningful spatial and non-spatial factors. This intuitive factorisation does not require the specific network architecture choices used here, but is general in nature and thus could be applied in diverse medical image analysis tasks.
Factorisation facilitates manipulations of the latent space and as such probing and interpreting the model. Such interpretability is considered key to advance the translation of advanced machine learning methods in the clinic (and perhaps why it has been recently emphasised with dedicated MICCAI workshops \url{http://interpretable-ml.org/miccai2018tutorial/}).

Our work has some limitations that inspire future directions. We can envision that extensions to 3D (in lieu of 2D), and the explicit learning of hierarchical factors that better capture semantic information (both in terms of anatomical and modality representations), would further improve applicability of our approach in several domains such as brain (which benefits from 3D view) and abdominal imaging. This work further encourages future extensions to improve the fidelity of reconstructed images by explicitly modelling image texture, which would benefit applications in ultrasound. 
This can be achieved with the design of more powerful decoders, although how best to maintain the balance between the semantics of the spatial representation and the quality of the reconstruction is an open question.
Finally, future work includes the extension of the method's applicability in a completely unsupervised setting where no annotated examples are available.

\section*{Acknowledgements}

This work was supported in part by the US National Institutes of Health (1R01HL136578-01) and UK EPSRC (EP/P022928/1), and used resources provided by the Edinburgh Compute and Data Facility (http://www.ecdf.ed.ac.uk/). S.A. Tsaftaris acknowledges the support of the Royal Academy of Engineering and the Research Chairs and Senior Research Fellowships scheme. 

\bibliographystyle{plainnat}
\bibliography{references}

\begin{thebibliography}{50}
\providecommand{\natexlab}[1]{#1}
\providecommand{\url}[1]{\texttt{#1}}
\expandafter\ifx\csname urlstyle\endcsname\relax
  \providecommand{\doi}[1]{doi: #1}\else
  \providecommand{\doi}{doi: \begingroup \urlstyle{rm}\Url}\fi

\bibitem[Almahairi et~al.(2018)Almahairi, Rajeswar, Sordoni, Bachman, and
  Courville]{almahairi2018augmented}
Amjad Almahairi, Sai Rajeswar, Alessandro Sordoni, Philip Bachman, and Aaron~C.
  Courville.
\newblock Augmented {CycleGAN}: Learning many-to-many mappings from unpaired
  data.
\newblock In \emph{International Conference on Machine Learning}, 2018.

\bibitem[Azadi et~al.(2018)Azadi, Fisher, Kim, Wang, Shechtman, and
  Darrell]{azadi2018multi}
Samaneh Azadi, Matthew Fisher, Vladimir Kim, Zhaowen Wang, Eli Shechtman, and
  Trevor Darrell.
\newblock Multi-content {GAN} for few-shot font style transfer.
\newblock In \emph{Proceedings of the IEEE Conference on Computer Vision and
  Pattern Recognition}, volume~11, page~13, 2018.

\bibitem[Bai et~al.(2017)Bai, Oktay, Sinclair, Suzuki, Rajchl, Tarroni,
  Glocker, King, Matthews, and Rueckert]{Bai2017}
Wenjia Bai, Ozan Oktay, Matthew Sinclair, Hideaki Suzuki, Martin Rajchl,
  Giacomo Tarroni, Ben Glocker, Andrew King, Paul~M Matthews, and Daniel
  Rueckert.
\newblock Semi-supervised learning for network-based cardiac {MR} image
  segmentation.
\newblock In \emph{Medical Image Computing and Computer-Assisted Intervention},
  pages 253--260, Cham, 2017. Springer International Publishing.
\newblock ISBN 978-3-319-66185-8.

\bibitem[Bai et~al.(2018{\natexlab{a}})Bai, Sinclair, Tarroni, Oktay, Rajchl,
  Vaillant, Lee, Aung, Lukaschuk, Sanghvi, Zemrak, Fung, Paiva, Carapella, Kim,
  Suzuki, Kainz, Matthews, Petersen, Piechnik, Neubauer, Glocker, and
  Rueckert]{Bai2018}
Wenjia Bai, Matthew Sinclair, Giacomo Tarroni, Ozan Oktay, Martin Rajchl,
  Ghislain Vaillant, Aaron~M. Lee, Nay Aung, Elena Lukaschuk, Mihir~M. Sanghvi,
  Filip Zemrak, Kenneth Fung, Jose~Miguel Paiva, Valentina Carapella, Young~Jin
  Kim, Hideaki Suzuki, Bernhard Kainz, Paul~M. Matthews, Steffen~E. Petersen,
  Stefan~K. Piechnik, Stefan Neubauer, Ben Glocker, and Daniel Rueckert.
\newblock Automated cardiovascular magnetic resonance image analysis with fully
  convolutional networks.
\newblock \emph{Journal of Cardiovascular Magnetic Resonance}, 20\penalty0
  (1):\penalty0 65, Sep 2018{\natexlab{a}}.
\newblock \doi{10.1186/s12968-018-0471-x}.

\bibitem[Bai et~al.(2018{\natexlab{b}})Bai, Suzuki, Qin, Tarroni, Oktay,
  Matthews, and Rueckert]{Bai2018_Recurrent}
Wenjia Bai, Hideaki Suzuki, Chen Qin, Giacomo Tarroni, Ozan Oktay, Paul~M.
  Matthews, and Daniel Rueckert.
\newblock Recurrent neural networks for aortic image sequence segmentation with
  sparse annotations.
\newblock In Alejandro~F. Frangi, Julia~A. Schnabel, Christos Davatzikos,
  Carlos Alberola-L{\'o}pez, and Gabor Fichtinger, editors, \emph{Medical Image
  Computing and Computer Assisted Intervention}, pages 586--594, Cham,
  2018{\natexlab{b}}. Springer International Publishing.
\newblock ISBN 978-3-030-00937-3.

\bibitem[Bengio et~al.(2013{\natexlab{a}})Bengio, Courville, and
  Vincent]{bengio2013representation}
Yoshua Bengio, Aaron Courville, and Pascal Vincent.
\newblock Representation learning: A review and new perspectives.
\newblock \emph{IEEE transactions on pattern analysis and machine
  intelligence}, 35\penalty0 (8):\penalty0 1798--1828, 2013{\natexlab{a}}.
\newblock \doi{10.1109/TPAMI.2013.50}.

\bibitem[Bengio et~al.(2013{\natexlab{b}})Bengio, Léonard, and
  Courville]{BengioLC13}
Yoshua Bengio, Nicholas Léonard, and Aaron~C. Courville.
\newblock Estimating or propagating gradients through stochastic neurons for
  conditional computation.
\newblock \emph{CoRR}, abs/1308.3432, 2013{\natexlab{b}}.

\bibitem[Bernard et~al.(2018)Bernard, Lalande, Zotti, Cervenansky, Yang, Heng,
  Cetin, Lekadir, Camara, Ballester, Sanroma, Napel, Petersen, Tziritas,
  Grinias, Khened, Kollerathu, Krishnamurthi, Rohé, Pennec, Sermesant,
  Isensee, Jäger, Maier-Hein, Baumgartner, Koch, Wolterink, Išgum, Jang,
  Hong, Patravali, Jain, Humbert, and Jodoin]{BernardACDC}
O.~Bernard, A.~Lalande, C.~Zotti, F.~Cervenansky, X.~Yang, P.~Heng, I.~Cetin,
  K.~Lekadir, O.~Camara, M.~A.~G. Ballester, G.~Sanroma, S.~Napel, S.~Petersen,
  G.~Tziritas, E.~Grinias, M.~Khened, V.~A. Kollerathu, G.~Krishnamurthi,
  M.~Rohé, X.~Pennec, M.~Sermesant, F.~Isensee, P.~Jäger, K.~H. Maier-Hein,
  C.~F. Baumgartner, L.~M. Koch, J.~M. Wolterink, I.~Išgum, Y.~Jang, Y.~Hong,
  J.~Patravali, S.~Jain, O.~Humbert, and P.~Jodoin.
\newblock Deep learning techniques for automatic {MRI} cardiac multi-structures
  segmentation and diagnosis: Is the problem solved?
\newblock \emph{IEEE Transactions on Medical Imaging}, 37\penalty0
  (11):\penalty0 2514--2525, Nov 2018.
\newblock ISSN 0278-0062.
\newblock \doi{10.1109/TMI.2018.2837502}.

\bibitem[Bevilacqua et~al.(2016)Bevilacqua, Dharmakumar, and
  Tsaftaris]{bevilacqua2016dictionary}
Marco Bevilacqua, Rohan Dharmakumar, and Sotirios~A Tsaftaris.
\newblock Dictionary-driven ischemia detection from cardiac phase-resolved
  myocardial {BOLD MRI} at rest.
\newblock \emph{IEEE Transactions on Medical Imaging}, 35\penalty0
  (1):\penalty0 282--293, Jan 2016.
\newblock ISSN 0278-0062.
\newblock \doi{10.1109/TMI.2015.2470075}.

\bibitem[Biffi et~al.(2018)Biffi, Oktay, Tarroni, Bai, De~Marvao, Doumou,
  Rajchl, Bedair, Prasad, Cook, O'Regan, and
  Rueckert]{Biffi2018_LearningInterpretable}
Carlo Biffi, Ozan Oktay, Giacomo Tarroni, Wenjia Bai, Antonio De~Marvao,
  Georgia Doumou, Martin Rajchl, Reem Bedair, Sanjay Prasad, Stuart Cook,
  Declan O'Regan, and Daniel Rueckert.
\newblock Learning interpretable anatomical features through deep generative
  models: Application to cardiac remodeling.
\newblock In Alejandro~F. Frangi, Julia~A. Schnabel, Christos Davatzikos,
  Carlos Alberola-L{\'o}pez, and Gabor Fichtinger, editors, \emph{Medical Image
  Computing and Computer Assisted Intervention}, pages 464--471, Cham, 2018.
  Springer International Publishing.
\newblock ISBN 978-3-030-00934-2.

\bibitem[Burgess et~al.(2018)Burgess, Higgins, Pal, Matthey, Watters,
  Desjardins, and Lerchner]{burgess2018understanding}
Christopher~P Burgess, Irina Higgins, Arka Pal, Loic Matthey, Nick Watters,
  Guillaume Desjardins, and Alexander Lerchner.
\newblock Understanding disentangling in $\beta$-vae.
\newblock \emph{NIPS Workshop on Learning Disentangled Representations}, 2018.

\bibitem[Chartsias et~al.(2017)Chartsias, Joyce, Dharmakumar, and
  Tsaftaris]{chartsias2017adversarial}
Agisilaos Chartsias, Thomas Joyce, Rohan Dharmakumar, and Sotirios~A Tsaftaris.
\newblock Adversarial image synthesis for unpaired multi-modal cardiac data.
\newblock In \emph{Simulation and Synthesis in Medical Imaging}, pages 3--13.
  Springer International Publishing, 2017.
\newblock ISBN 978-3-319-68127-6.

\bibitem[Chartsias et~al.(2018)Chartsias, Joyce, Papanastasiou, Semple,
  Williams, Newby, Dharmakumar, and Tsaftaris]{chartsias2018factorised}
Agisilaos Chartsias, Thomas Joyce, Giorgos Papanastasiou, Scott Semple,
  Michelle Williams, David Newby, Rohan Dharmakumar, and Sotirios~A. Tsaftaris.
\newblock Factorised spatial representation learning: Application in
  semi-supervised myocardial segmentation.
\newblock In \emph{Medical Image Computing and Computer Assisted Intervention},
  pages 490--498, Cham, 2018. Springer International Publishing.
\newblock ISBN 978-3-030-00934-2.

\bibitem[Chen et~al.(2016)Chen, Duan, Houthooft, Schulman, Sutskever, and
  Abbeel]{chen2016infogan}
Xi~Chen, Yan Duan, Rein Houthooft, John Schulman, Ilya Sutskever, and Pieter
  Abbeel.
\newblock Info{GAN}: Interpretable representation learning by information
  maximizing generative adversarial nets.
\newblock In \emph{Advances in neural information processing systems}, pages
  2172--2180. Curran Associates, Inc., 2016.

\bibitem[Cheplygina et~al.(2018)Cheplygina, de~Bruijne, and
  Pluim]{cheplygina2018not}
Veronika Cheplygina, Marleen de~Bruijne, and Josien P.~W. Pluim.
\newblock Not-so-supervised: a survey of semi-supervised, multi-instance, and
  transfer learning in medical image analysis.
\newblock \emph{CoRR}, abs/1804.06353, 2018.

\bibitem[Chollet et~al.(2015)]{chollet2015keras}
Fran\c{c}ois Chollet et~al.
\newblock Keras.
\newblock \url{https://keras.io}, 2015.

\bibitem[Donahue et~al.(2018)Donahue, Lipton, Balsubramani, and
  McAuley]{donahue2017semantically}
Chris Donahue, Zachary~C Lipton, Akshay Balsubramani, and Julian McAuley.
\newblock Semantically decomposing the latent spaces of generative adversarial
  networks.
\newblock In \emph{International Conference on Learning Representations}, 2018.

\bibitem[Esser et~al.(2018)Esser, Sutter, and Ommer]{esser2018variational}
Patrick Esser, Ekaterina Sutter, and Bj{\"o}rn Ommer.
\newblock A variational u-net for conditional appearance and shape generation.
\newblock In \emph{Proceedings of the IEEE Conference on Computer Vision and
  Pattern Recognition}, pages 8857--8866, 2018.

\bibitem[Fidon et~al.(2017)Fidon, Li, Garcia-Peraza-Herrera, Ekanayake,
  Kitchen, Ourselin, and Vercauteren]{fidon2017scalable}
Lucas Fidon, Wenqi Li, Luis~C Garcia-Peraza-Herrera, Jinendra Ekanayake, Neil
  Kitchen, Sebastien Ourselin, and Tom Vercauteren.
\newblock Scalable multimodal convolutional networks for brain tumour
  segmentation.
\newblock In \emph{Medical Image Computing and Computer-Assisted Intervention},
  pages 285--293, Cham, 2017. Springer International Publishing.
\newblock ISBN 978-3-319-66179-7.

\bibitem[Gatys et~al.(2016)Gatys, Ecker, and Bethge]{gatys2016image}
Leon~A Gatys, Alexander~S Ecker, and Matthias Bethge.
\newblock Image style transfer using convolutional neural networks.
\newblock In \emph{Proceedings of the IEEE Conference on Computer Vision and
  Pattern Recognition}, pages 2414--2423, 2016.
\newblock \doi{10.1109/CVPR.2016.265}.

\bibitem[Goodfellow et~al.(2014)Goodfellow, Pouget-Abadie, Mirza, Xu,
  Warde-Farley, Ozair, Courville, and Bengio]{goodfellow2014generative}
Ian Goodfellow, Jean Pouget-Abadie, Mehdi Mirza, Bing Xu, David Warde-Farley,
  Sherjil Ozair, Aaron Courville, and Yoshua Bengio.
\newblock Generative adversarial nets.
\newblock In \emph{Advances in neural information processing systems}, pages
  2672--2680. Curran Associates, Inc., 2014.

\bibitem[Higgins et~al.(2017)Higgins, Matthey, Pal, Burgess, Glorot, Botvinick,
  Mohamed, and Lerchner]{higgins2016beta}
Irina Higgins, Loic Matthey, Arka Pal, Christopher Burgess, Xavier Glorot,
  Matthew Botvinick, Shakir Mohamed, and Alexander Lerchner.
\newblock beta-vae: Learning basic visual concepts with a constrained
  variational framework.
\newblock In \emph{International Conference on Learning Representations}, 2017.

\bibitem[Hu et~al.(2017)Hu, Szab{\'o}, Portenier, Zwicker, and
  Favaro]{hu2017disentangling}
Qiyang Hu, Attila Szab{\'o}, Tiziano Portenier, Matthias Zwicker, and Paolo
  Favaro.
\newblock Disentangling factors of variation by mixing them.
\newblock \emph{CoRR}, abs/1711.07410, 2017.

\bibitem[Huang et~al.(2018)Huang, Liu, Belongie, and Kautz]{huang2018munit}
Xun Huang, Ming-Yu Liu, Serge Belongie, and Jan Kautz.
\newblock Multimodal unsupervised image-to-image translation.
\newblock In \emph{European Conference on Computer Vision}, volume 11207, pages
  179--196. Springer International Publishing, 2018.

\bibitem[Kim and Mnih(2018)]{kim2018disentangling}
Hyunjik Kim and Andriy Mnih.
\newblock Disentangling by factorising.
\newblock In \emph{International Conference on Machine Learning}, volume~80 of
  \emph{{JMLR} Workshop and Conference Proceedings}, pages 2654--2663.
  JMLR.org, 2018.

\bibitem[Kingma and Ba(2014)]{KingmaB14}
Diederik~P. Kingma and Jimmy Ba.
\newblock Adam: {A} method for stochastic optimization.
\newblock \emph{CoRR}, abs/1412.6980, 2014.

\bibitem[Kingma and Welling(2014)]{kingma2013auto}
Diederik~P. Kingma and Max Welling.
\newblock Auto-encoding variational bayes.
\newblock In \emph{International Conference on Learning Representations}, 2014.

\bibitem[Lee et~al.(2018)Lee, Tseng, Huang, Singh, and Yang]{lee2018diverse}
Hsin-Ying Lee, Hung-Yu Tseng, Jia-Bin Huang, Maneesh~Kumar Singh, and
  Ming-Hsuan Yang.
\newblock Diverse image-to-image translation via disentangled representations.
\newblock In \emph{European Conference on Computer Vision}, volume 11205, pages
  36--52. Springer International Publishing, 2018.

\bibitem[Mao et~al.(2018)Mao, Li, Xie, Lau, Wang, and
  Smolley]{mao2017effectiveness}
Xudong Mao, Qing Li, Haoran Xie, Raymond Y.~K. Lau, Zhen Wang, and Stephen~Paul
  Smolley.
\newblock On the effectiveness of least squares generative adversarial
  networks.
\newblock \emph{IEEE transactions on pattern analysis and machine
  intelligence}, 2018.
\newblock ISSN 0162-8828.
\newblock \doi{10.1109/TPAMI.2018.2872043}.

\bibitem[Mathieu et~al.(2016)Mathieu, Zhao, Zhao, Ramesh, Sprechmann, and
  LeCun]{mathieu2016disentangling}
Michael~F Mathieu, Junbo~Jake Zhao, Junbo Zhao, Aditya Ramesh, Pablo
  Sprechmann, and Yann LeCun.
\newblock Disentangling factors of variation in deep representation using
  adversarial training.
\newblock In \emph{Advances in Neural Information Processing Systems}, pages
  5040--5048, 2016.

\bibitem[Milletari et~al.(2016)Milletari, Navab, and
  Ahmadi]{Milletari2016VNetFC}
Fausto Milletari, Nassir Navab, and Seyed-Ahmad Ahmadi.
\newblock {V-Net}: Fully convolutional neural networks for volumetric medical
  image segmentation.
\newblock \emph{2016 Fourth International Conference on 3D Vision}, pages
  565--571, 2016.
\newblock \doi{10.1109/3DV.2016.79}.

\bibitem[Nie et~al.(2018)Nie, Gao, Wang, and Shen]{nie2018asdnet}
Dong Nie, Yaozong Gao, Li~Wang, and Dinggang Shen.
\newblock {ASDNet}: Attention based semi-supervised deep networks for medical
  image segmentation.
\newblock In \emph{Medical Image Computing and Computer Assisted Intervention},
  pages 370--378, Cham, 2018. Springer International Publishing.

\bibitem[Oktay et~al.(2018)Oktay, Ferrante, Kamnitsas, Heinrich, Bai,
  Caballero, Cook, de~Marvao, Dawes, O‘Regan, Kainz, Glocker, and
  Rueckert]{OktayAnatomically}
O.~Oktay, E.~Ferrante, K.~Kamnitsas, M.~Heinrich, W.~Bai, J.~Caballero, S.~A.
  Cook, A.~de~Marvao, T.~Dawes, D.~P. O‘Regan, B.~Kainz, B.~Glocker, and
  D.~Rueckert.
\newblock Anatomically constrained neural networks (acnns): Application to
  cardiac image enhancement and segmentation.
\newblock \emph{IEEE Transactions on Medical Imaging}, 37\penalty0
  (2):\penalty0 384--395, Feb 2018.
\newblock ISSN 0278-0062.
\newblock \doi{10.1109/TMI.2017.2743464}.

\bibitem[Peng et~al.(2016)Peng, Lekadir, Gooya, Shao, Petersen, and
  Frangi]{peng2016review}
Peng Peng, Karim Lekadir, Ali Gooya, Ling Shao, Steffen~E Petersen, and
  Alejandro~F Frangi.
\newblock A review of heart chamber segmentation for structural and functional
  analysis using cardiac magnetic resonance imaging.
\newblock \emph{Magnetic Resonance Materials in Physics, Biology and Medicine},
  29\penalty0 (2):\penalty0 155--195, 2016.

\bibitem[Perez et~al.(2018)Perez, Strub, de~Vries, Dumoulin, and
  Courville]{perez2017film}
Ethan Perez, Florian Strub, Harm de~Vries, Vincent Dumoulin, and Aaron~C.
  Courville.
\newblock {FiLM}: Visual reasoning with a general conditioning layer.
\newblock In \emph{{AAAI}}, pages 3942--3951. {AAAI} Press, 2018.

\bibitem[Qin et~al.(2018)Qin, Bai, Schlemper, Petersen, Piechnik, Neubauer, and
  Rueckert]{Qin2018_JointMotion}
Chen Qin, Wenjia Bai, Jo~Schlemper, Steffen~E. Petersen, Stefan~K. Piechnik,
  Stefan Neubauer, and Daniel Rueckert.
\newblock Joint motion estimation and segmentation from undersampled cardiac mr
  image.
\newblock In Florian Knoll, Andreas Maier, and Daniel Rueckert, editors,
  \emph{Machine Learning for Medical Image Reconstruction}, pages 55--63, Cham,
  2018. Springer International Publishing.
\newblock ISBN 978-3-030-00129-2.

\bibitem[Radford et~al.(2015)Radford, Metz, and
  Chintala]{Radford2015UnsupervisedRL}
Alec Radford, Luke Metz, and Soumith Chintala.
\newblock Unsupervised representation learning with deep convolutional
  generative adversarial networks.
\newblock \emph{CoRR}, abs/1511.06434, 2015.

\bibitem[Rezende et~al.(2014)Rezende, Mohamed, and
  Wierstra]{rezende2014stochastic}
Danilo~Jimenez Rezende, Shakir Mohamed, and Daan Wierstra.
\newblock Stochastic backpropagation and approximate inference in deep
  generative models.
\newblock In \emph{International Conference on Machine Learning}, volume~32 of
  \emph{Proceedings of Machine Learning Research}, pages 1278--1286. PMLR,
  22--24 Jun 2014.

\bibitem[Ronneberger et~al.(2015)Ronneberger, Fischer, and
  Brox]{ronneberger2015u}
Olaf Ronneberger, Philipp Fischer, and Thomas Brox.
\newblock {U-Net}: Convolutional networks for biomedical image segmentation.
\newblock In \emph{Medical Image Computing and Computer-Assisted Intervention},
  pages 234--241, Cham, 2015. Springer International Publishing.

\bibitem[Szab{\'o} et~al.(2018)Szab{\'o}, Hu, Portenier, Zwicker, and
  Favaro]{szabo2017challenges}
Attila Szab{\'o}, Qiyang Hu, Tiziano Portenier, Matthias Zwicker, and Paolo
  Favaro.
\newblock Challenges in disentangling independent factors of variation.
\newblock In \emph{International Conference on Learning Representations
  Workshop}, 2018.

\bibitem[Tan et~al.(2017)Tan, Liew, Lim, and McLaughlin]{TAN201778}
Li~Kuo Tan, Yih~Miin Liew, Einly Lim, and Robert~A. McLaughlin.
\newblock Convolutional neural network regression for short-axis left ventricle
  segmentation in cardiac cine {MR} sequences.
\newblock \emph{Medical Image Analysis}, 39:\penalty0 78 -- 86, 2017.
\newblock ISSN 1361-8415.
\newblock \doi{https://doi.org/10.1016/j.media.2017.04.002}.

\bibitem[Tsaftaris et~al.(2013)Tsaftaris, Zhou, Tang, Li, and
  Dharmakumar]{tsaftaris2013detecting}
Sotirios~A Tsaftaris, Xiangzhi Zhou, Richard Tang, Debiao Li, and Rohan
  Dharmakumar.
\newblock Detecting myocardial ischemia at rest with cardiac phase--resolved
  blood oxygen level--dependent cardiovascular magnetic resonance.
\newblock \emph{Circulation: Cardiovascular Imaging}, 6\penalty0 (2):\penalty0
  311--319, 2013.

\bibitem[Vigneault et~al.(2018)Vigneault, Xie, Ho, Bluemke, and
  Noble]{VIGNEAULT201895}
Davis~M. Vigneault, Weidi Xie, Carolyn~Y. Ho, David~A. Bluemke, and J.~Alison
  Noble.
\newblock ${\Omega} $-net (omega-net): Fully automatic, multi-view cardiac mr
  detection, orientation, and segmentation with deep neural networks.
\newblock \emph{Medical Image Analysis}, 48:\penalty0 95 -- 106, 2018.
\newblock ISSN 1361-8415.
\newblock \doi{https://doi.org/10.1016/j.media.2018.05.008}.

\bibitem[Zhang et~al.(2017)Zhang, Yang, Chen, Fredericksen, Hughes, and
  Chen]{ZhangYizhe2017}
Yizhe Zhang, Lin Yang, Jianxu Chen, Maridel Fredericksen, David~P Hughes, and
  Danny~Z Chen.
\newblock Deep adversarial networks for biomedical image segmentation utilizing
  unannotated images.
\newblock In \emph{Medical Image Computing and Computer-Assisted Intervention},
  pages 408--416, Cham, 2017. Springer International Publishing.

\bibitem[Zhao et~al.(2018)Zhao, Yang, Zheng, Guldner, Zhang, and
  Chen]{zhao2018deep}
Zhuo Zhao, Lin Yang, Hao Zheng, Ian~H Guldner, Siyuan Zhang, and Danny~Z Chen.
\newblock Deep learning based instance segmentation in {3D} biomedical images
  using weak annotation.
\newblock In \emph{Medical Image Computing and Computer Assisted Intervention},
  pages 352--360, Cham, 2018. Springer International Publishing.

\bibitem[Zheng et~al.(2018)Zheng, Delingette, Duchateau, and
  Ayache]{Zheng3DConsistent}
Q.~Zheng, H.~Delingette, N.~Duchateau, and N.~Ayache.
\newblock {3-D} consistent and robust segmentation of cardiac images by deep
  learning with spatial propagation.
\newblock \emph{IEEE Transactions on Medical Imaging}, 37\penalty0
  (9):\penalty0 2137--2148, Sept 2018.
\newblock ISSN 0278-0062.
\newblock \doi{10.1109/TMI.2018.2820742}.

\bibitem[Zhu et~al.(2017)Zhu, Zhang, Pathak, Darrell, Efros, Wang, and
  Shechtman]{zhu2017toward}
Jun-Yan Zhu, Richard Zhang, Deepak Pathak, Trevor Darrell, Alexei~A Efros,
  Oliver Wang, and Eli Shechtman.
\newblock Toward multimodal image-to-image translation.
\newblock In \emph{Advances in Neural Information Processing Systems}, pages
  465--476, 2017.

\bibitem[Zhuang(2013)]{zhuang2013challenges}
Xiahai Zhuang.
\newblock Challenges and methodologies of fully automatic whole heart
  segmentation: a review.
\newblock \emph{Journal of Healthcare Engineering}, 4\penalty0 (3):\penalty0
  371--407, 2013.

\bibitem[Zhuang and Shen(2016)]{zhuang2016multi}
Xiahai Zhuang and Juan Shen.
\newblock Multi-scale patch and multi-modality atlases for whole heart
  segmentation of {MRI}.
\newblock \emph{Medical Image Analysis}, 31:\penalty0 77--87, 2016.
\newblock \doi{https://doi.org/10.1016/j.media.2016.02.006}.

\bibitem[Zhuang et~al.(2010)Zhuang, Rhode, Razavi, Hawkes, and
  Ourselin]{zhuang2010registration}
Xiahai Zhuang, Kawal~S Rhode, Reza~S Razavi, David~J Hawkes, and Sebastien
  Ourselin.
\newblock A registration-based propagation framework for automatic whole heart
  segmentation of cardiac {MRI}.
\newblock \emph{IEEE Transactions on Medical Imaging}, 29\penalty0
  (9):\penalty0 1612--25, 2010.
\newblock \doi{10.1109/TMI.2010.2047112}.

\end{thebibliography}

\end{document}